\documentclass[10pt,twocolumn,letterpaper]{article}

\usepackage{iccv}
\usepackage{times}
\usepackage{epsfig}
\usepackage{graphicx}
\usepackage{amsmath}
\usepackage{amssymb}
\usepackage{makecell}
\usepackage{gensymb}

\usepackage{authblk}
\usepackage{multirow}
\usepackage[caption=false]{subfig}

\usepackage[breaklinks=true,bookmarks=false]{hyperref}

\iccvfinalcopy 

\def\iccvPaperID{2986} 

\ificcvfinal\fi

\begin{document}

\title{Revisiting Stereo Depth Estimation From a Sequence-to-Sequence Perspective with Transformers}

\author{Zhaoshuo Li}
\author{Xingtong Liu}
\author{Nathan Drenkow}
\author{Andy Ding}
\author{Francis X. Creighton}
\author{Russell H. Taylor}
\author{Mathias Unberath}
\affil{Johns Hopkins University \protect\\ \tt\small \{zli122, mathias\}@jhu.edu}

\maketitle
\ificcvfinal\thispagestyle{empty}\fi

\begin{abstract}
   Stereo depth estimation relies on optimal correspondence matching between pixels on epipolar lines in the left and right images to infer depth. In this work, we revisit the problem from a sequence-to-sequence correspondence perspective to replace cost volume construction with dense pixel matching using position information and attention. This approach, named STereo TRansformer (STTR), has several advantages: It 1) relaxes the limitation of a fixed disparity range, 2) identifies occluded regions and provides confidence estimates, and 3) imposes uniqueness constraints during the matching process. We report promising results on both synthetic and real-world datasets and demonstrate that STTR generalizes across different domains, even without fine-tuning. 
\end{abstract}


\section{Introduction}

Stereo depth estimation is of substantial interest since it enables the reconstruction of 3D information. To this end, corresponding pixels are matched between the left and right camera image; the difference in corresponding pixel location, i.\,e. the disparity, can then be used to infer depth and reconstruct the 3D scene. Recent deep learning-based approaches to stereo depth estimation have shown promising results but several challenges remain. 

One such challenge relates to the use of a limited disparity range. Disparity values can, in theory, range from zero to the image width depending on the resolution/baseline of the cameras, and their proximity to the physical objects. However, many of the best performing approaches are constrained to a manually pre-specified disparity range (typically a maximum of 192 px) \cite{laga2020survey}. These methods rely on ``cost volumes" in which matching costs are computed for multiple candidate matches and a final predicted disparity value is computed as the aggregated sum. This self-imposed disparity range is necessary to enable memory-feasible implementations of these methods but is not flexible to properties of the physical scene and/or the camera setup. In applications such as autonomous driving and endoscopic intervention, it is important to recognize close objects irrespective of camera setup (with disparity values potentially larger than 192) to avoid collisions, suggesting the need to relax the fixed disparity range assumption.

Geometric properties and constraints such as occlusion and matching uniqueness, which led to the success of non-learning based approaches such as \cite{hirschmuller2007stereo}, are also often missing from learning-based approaches. For stereo depth estimation, occluded regions do not have a valid disparity. Prior algorithms generally infer disparities for occluded regions via a piece-wise smoothness assumption, which may not always be valid. Providing a confidence estimate together with the disparity value would be advantageous for down-stream analysis, such as for registration or scene understanding algorithms, to enable weighting or rejection of occluded and low-confidence estimates. However, most prior approaches do not provide such information. Moreover, pixels in one image should not be matched to multiple pixels in the other image (up to image resolution) since they correspond to the same location in the physical scene \cite{ohta1985stereo}. Although this constraint can be clearly useful to resolve ambiguity, most existing learning-based approaches do not impose it.

\begin{figure*}[htpb]
    \centering
    \subfloat[Network overview]{\includegraphics[width=0.6\linewidth, trim={4cm 7cm 3.4cm 5cm},clip]{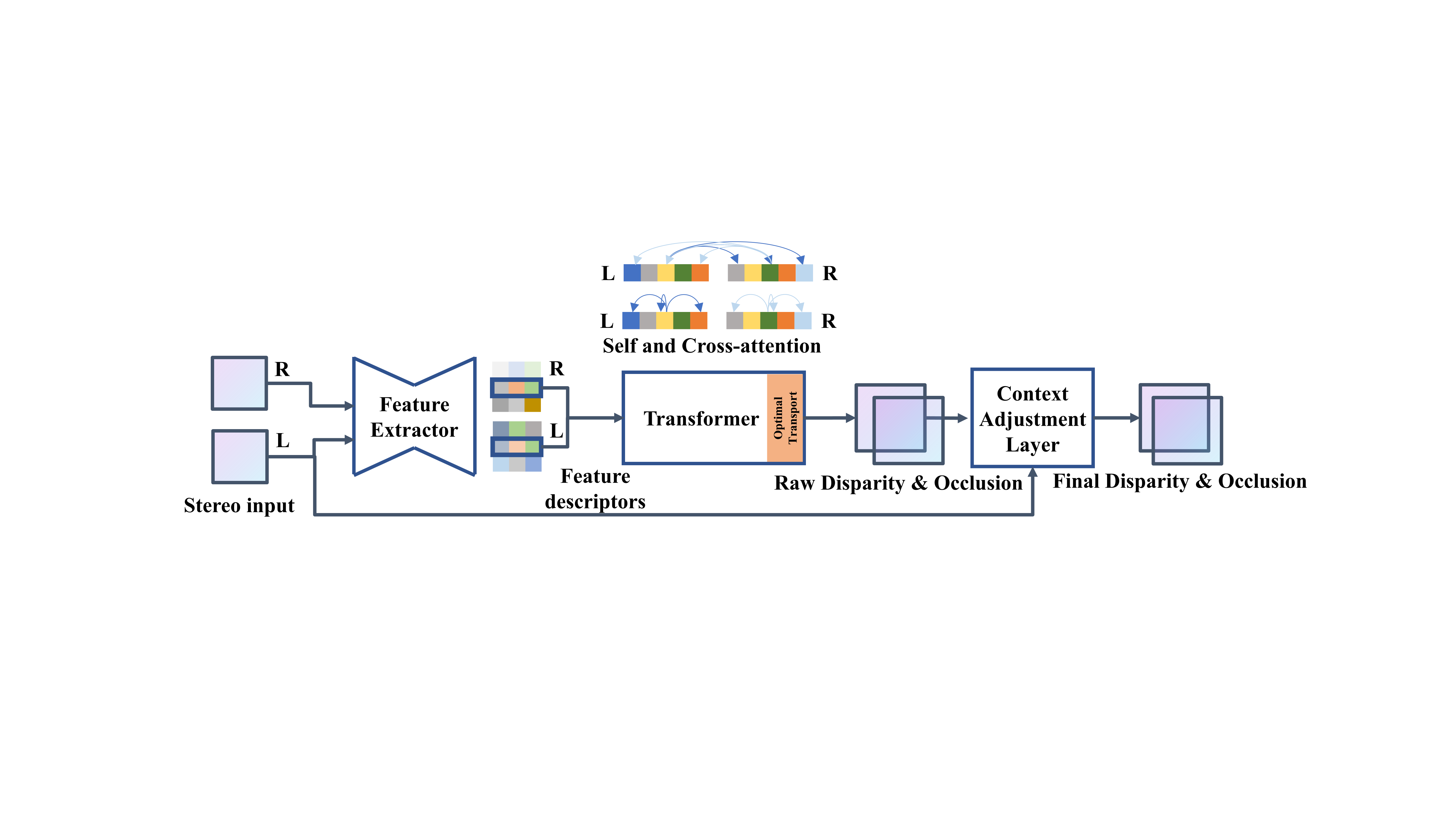}}
    
    \subfloat[Scene Flow]{\includegraphics[height=0.12\linewidth,trim={5cm 2cm 5cm 2cm},clip]{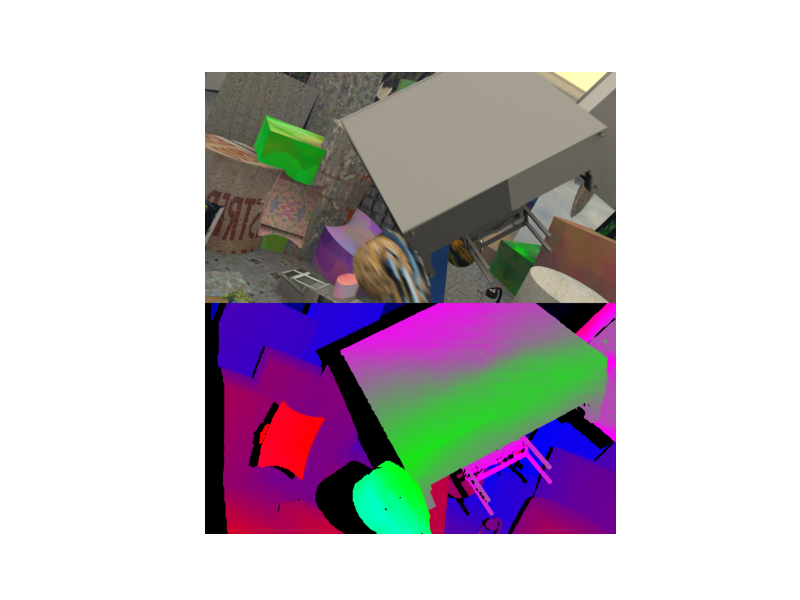}}
    \hspace{0.7em}
    \subfloat[MPI Sintel]{\includegraphics[height=0.12\linewidth,trim={0cm 0.5cm 1cm 1cm},clip]{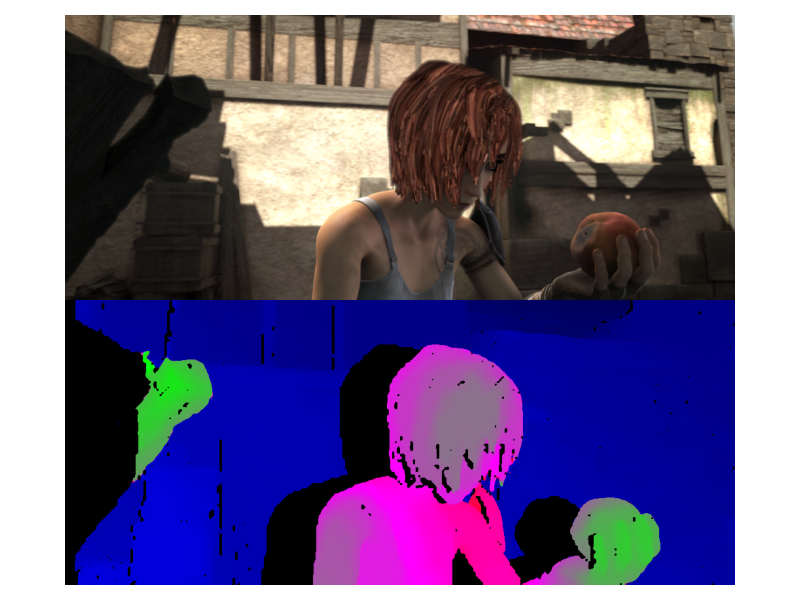}}
    \hspace{0.7em}
    \subfloat[KITTI 2015]{\includegraphics[height=0.12\linewidth,trim={2cm 3cm 4cm 2.5cm},clip]{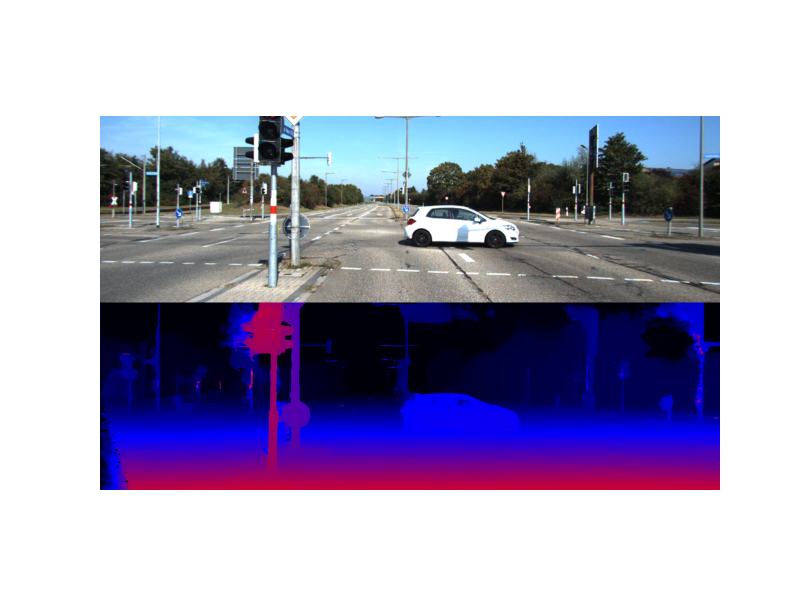}}
    \hspace{0.5em}
    \subfloat[Middlebury 2014]{\includegraphics[height=0.12\linewidth,trim={4cm 2cm 3cm 1.5cm},clip]{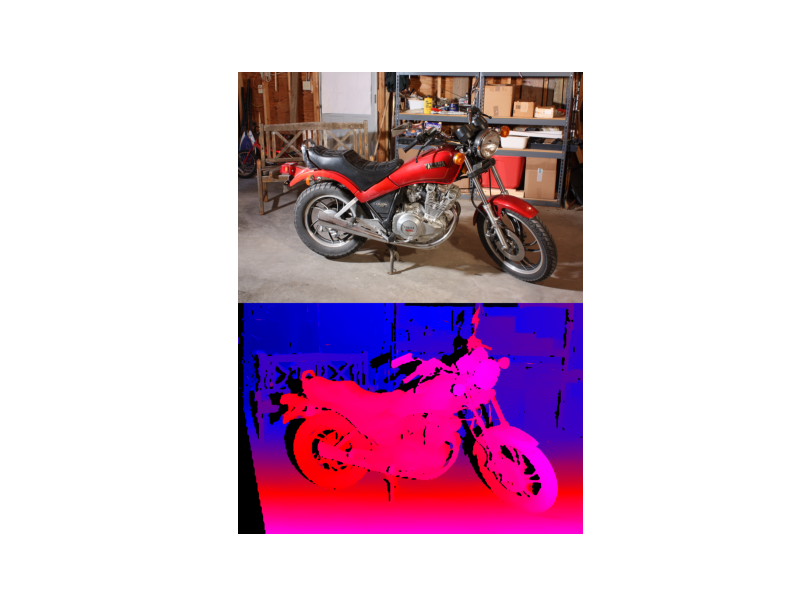}}
    \subfloat[SCARED]{\includegraphics[height=0.12\linewidth,trim={4cm 2cm 6cm 2cm},clip]{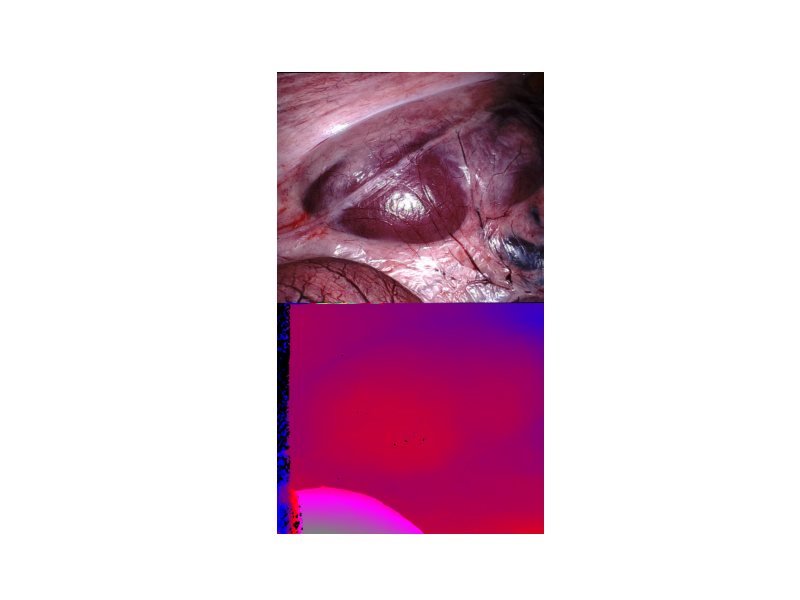}}
    \subfloat{\includegraphics[height=0.12\linewidth,trim={11.5cm 1cm 0cm 1cm},clip]{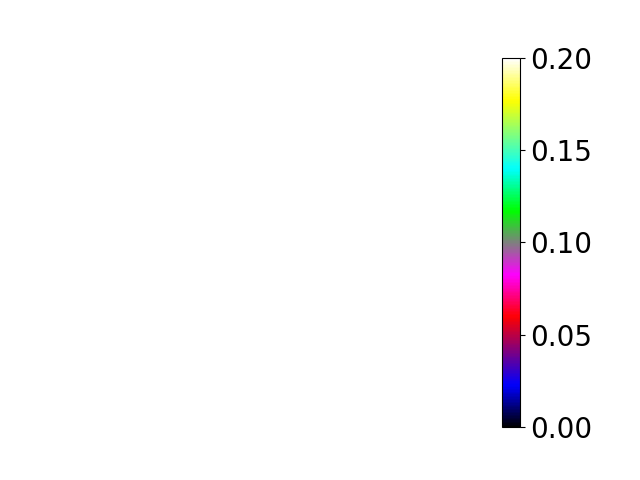}}
    
    \caption{(a) STTR estimates disparity by first extracting features from stereo images using a shared feature extractor. The extracted feature descriptors are then used by a Transformer for dense self- and cross-attention computation, yielding a raw disparity estimate. A context adjustment layer further refines the disparity with information across epipolar lines conditioned on the left image for cross epipolar line optimality. (b-f) Inference of STTR trained only on synthetic Scene Flow dataset. Top row shows the left images. Bottom row shows predicted disparities. The color map used to visualize disparity is relative to the image width and is shown on the right. Black color indicates occlusion. Best viewed in color.}
    \label{fig:overview}
\end{figure*}

The aforementioned problems largely arise from shortcomings of the contemporary view of stereo matching which attempts to construct a cost volume. Approaches that consider disparity estimation from a sequence-to-sequence matching perspective along epipolar lines can avoid these challenges. Such methods are not new, to our knowledge, the first attempt using dynamic programming was proposed in 1985 \cite{ohta1985stereo}, where intra- and inter-epipolar line information is used together with a uniqueness constraint. However, it only used similarities between pixel intensities as matching criteria, which is inadequate beyond local matching, and thus restricted its performance. Recent advances in attention-based networks that capture long-range associations between feature descriptors prompt us to revisit this perspective. We take advantage of the recent Transformer architecture \cite{vaswani2017attention} proposed for language processing and recent advances in feature matching \cite{sarlin2020superglue}, and present a new end-to-end-trained stereo depth estimation network named STereo TRansformer (STTR). The main advantage of STTR is that it computes pixel-wise correlation densely and does not construct a fixed-disparity cost volume. Therefore, STTR can mitigate the drawbacks of most contemporary approaches that were detailed above with little to no compromises in performance. We report competitive performance on synthetic and real image benchmarks and demonstrate that STTR trained only on synthetic images also generalizes well to other domains without refinement. 

We make the following technical advances to enable the realization of STTR:
\begin{itemize}
    \itemsep0em
    \item[--] Instead of pixel-wise intensity correlation in traditional stereo depth estimation methods like \cite{ohta1985stereo}, we adopt a Transformer with alternating self- and cross-attentions combined with the optimal transport theory previously demonstrated in sparse feature matching \cite{sarlin2020superglue}. This design allows us to match pixels explicitly and densely while imposing a uniqueness constraint.
    \item[--] We provide a relative pixel distance encoding to the feature descriptor and use a customized attention mechanism to define discriminative features during the matching process. This helps resolve ambiguity during the matching process. 
    \item[--] 
    We devise a memory-feasible implementation of STTR which enables the training of the proposed model on conventional hardware. For seamless distribution and reproducibility, our code is available online\footnote{Code is available at \href{https://github.com/mli0603/stereo-transformer}{\textit{https://github.com/mli0603/stereo-transformer}}} and only uses existing PyTorch functions \cite{paszke2019pytorch}.
\end{itemize}

\section{Related Work}
\subsection{Stereo Depth Estimation}
In general, the stereo depth estimation task involves two key steps (1) feature matching and (2) matching cost aggregation \cite{laga2020survey}. Traditionally, the task is solved by dynamic programming techniques where matchings are computed using pixel intensities and costs are aggregated either horizontally in 1D \cite{ohta1985stereo} or multi-directionally in 2D \cite{hirschmuller2007stereo}.

More recently, learning-based methods that match features by dot-product \textit{correlation} have emerged. Early ones such as \cite{chen2015deep} compute feature similarities based on a patch of features and refine the matching using a Markov Random Field. Subsequent approaches \cite{mayer2016large,liang2018learning} take advantage of learning-based feature extractors and compute similarities between the feature descriptors for each pixel. \cite{xu2020aanet} further advances the performance with cross-scale information aggregation.

Concurrently, networks such as \cite{kendall2017end,chang2018pyramid} build a 4D feature volume by concatenating features at different disparities and learn to compute/aggregate the matching cost by \textit{3D convolutions}. In \cite{zhang2019ga}, additional semi-global and local cost aggregation layers are proposed to improve the performance. Following the same idea of computing matching cost by learnt 3D convolutions, other works \cite{yang2019hierarchical,cheng2020hierarchical,yang2020cost,gu2020cascade} attempt to enable high resolution inference, mitigate the memory constraint, and/or leverage richer context information via a multi-resolution approach. 

Hybrid approaches like \cite{guo2019group} have also emerged, which combine explicit correlation and 3D convolutions for matching and cost aggregation respectively. Other works follow different design concepts, such as \cite{badki2020bi3d} with a classification-based approach for disparity estimation.

However, none of the above prior work exploits the sequential nature and geometric properties of stereo matching, which led to the success of non-learning based works such as \cite{ohta1985stereo,hirschmuller2007stereo}. Moreover, whether matching is computed by correlation or learnt by 3D convolutions, a maximum disparity is set to mitigate memory and computation demands in the above works. For each pixel, there is a fixed and finite set of discretized locations where a pixel can be mapped, thus generating a matching cost volume. For disparities beyond this pre-defined range, these  approaches simply cannot infer the correct match. This limits the generalization of the networks across different scenes and stereo camera configurations. In addition, most learning-based approaches do not handle occlusion explicitly, even if the disparity in occluded regions can theoretically be arbitrary. Lastly, no explicit uniqueness constraint is imposed during the matching process, which can inhibit performance due to inconsistencies of matching. 

\subsection{Comparison of STTR to Previous Learning-based Stereo Paradigms}
We use a convolutional neural network as the feature extractor that feeds into a Transformer to capture long-range associations between pixels. STTR exploits the sequential nature and geometric properties of stereo matching.

\textbf{STTR vs. Correlation-based Networks}: STTR imposes a uniqueness constraint during the matching process to resolve ambiguities. STTR also alternates between intra- and inter-image correlation operations, named self- and cross-attention, and updates the feature representations by considering both image context and position information.

\textbf{STTR vs. 3D Convolution-based Networks}: STTR explicitly and densely computes the correlation between pixels in the left and right images. Instead of using 3D convolutions to aggregate a cost volume, we first match along epipolar lines and then aggregate the information across epipolar lines via 2D convolutions.

\subsection{Attention Mechanism and Transformer}
Attention has already proven to be an effective tool in natural language processing \cite{vaswani2017attention}. Recently, attention-based architectures has found applications in computer vision tasks, such as image classification \cite{dosovitskiy2020image}, object detection \cite{carion2020end}, panoptic segmentation \cite{wang2020axial}, and homography estimation and visual localization \cite{sarlin2020superglue}, improving on the results of pure CNN architectures. This is likely because attention can capture long-range associations which is of particular importance for the work presented here. We adopt a Transformer to revisit the sequence-to-sequence stereo matching paradigm originally proposed in \cite{ohta1985stereo}.

\section{The Stereo Transformer Architecture}
In the following sections, we denote the height and width of the rectified left and right pair of images as $I_h$ and $I_w$. We denote the channel dimension of feature descriptors as $C$.
\subsection{Feature Extractor}
We use an hourglass-shaped architecture similar to \cite{liu2020extremely}, with the exception that the encoding path is modified with residual connections and spatial pyramid pooling modules \cite{chang2018pyramid} for more efficient global context acquisition. The decoding path consists of transposed convolution, dense-blocks \cite{huang2017densely}, and a final convolution layer. The feature descriptors for each pixel, denoted as vector $e_I$ of size $C_{e}$, encode both local and global context. The final feature map is at the same spatial resolution as the input image.

\subsection{Transformer}
\begin{figure}
    \centering
    \includegraphics[width=0.8\linewidth,trim={5.5cm 0cm 9cm 4cm},clip]{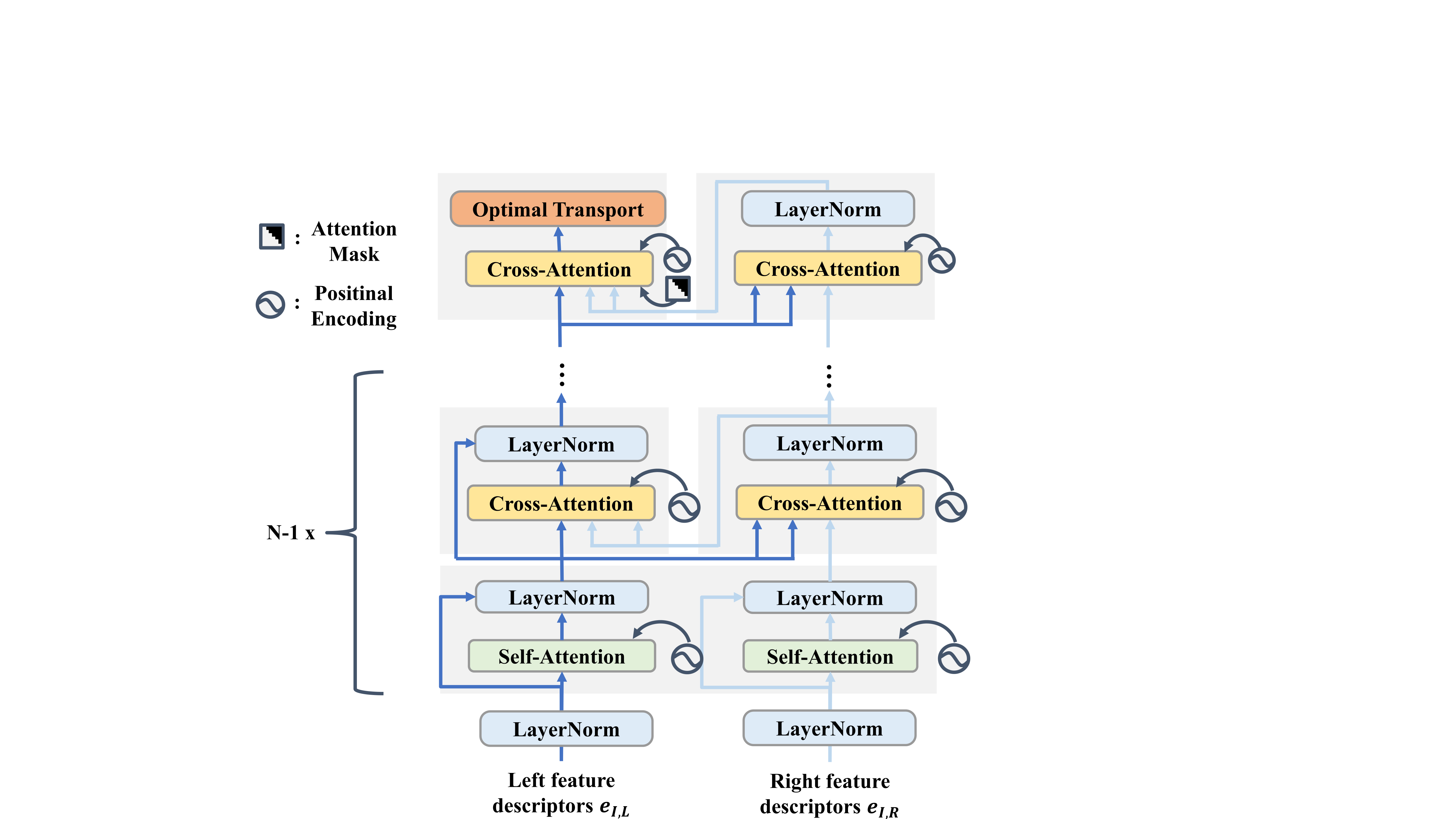}
    \caption{Overview of the Transformer module with alternating self- and cross-attention. Note that in the last cross-attention layer, the optimal transport and attention mask are added.}
    \label{fig:transformer_overview}
\end{figure}

An overview of the Transformer architecture used here is provided in \autoref{fig:transformer_overview}. We adopt the alternating attention mechanism in \cite{sarlin2020superglue}: \textit{Self-attention} computes attention between pixels along the epipolar line in the same image, while \textit{cross-attention} computes attention of pixels along corresponding epipolar lines in the left and right images. Details on both attention modules follows in \autoref{sssec:attention}. As shown in \autoref{fig:transformer_overview}, we alternate between computing self- and cross-attention for $N-1$ layers. This alternating scheme keeps updating the feature descriptors based on the image context and relative position, as discussed in \autoref{sssec:rel_pos_enc}.

In the last cross-attention layer, we use the most attended pixel to estimate the raw disparity. We add operations exclusive to this layer, including \textit{optimal transport} for compliance with the uniqueness constraint (\autoref{sssec:optimal_transport}) and an \textit{attention mask} for search space reduction (\autoref{sssec:attn_mask}).

\subsubsection{Attention}
\label{sssec:attention}
Attention modules \cite{vaswani2017attention} compute the attention between a set of \textit{query} vectors and \textit{key} vectors using dot-product similarity, which is then used to weigh a set of \textit{value} vectors.

We adopt multi-head attention, which increases the expressivity of the feature descriptor by splitting the channel dimension of feature descriptors $C_e$ into $N_h$ groups $C_{h}=C_{e}/N_{h}$, where $C_h$ is the channel dimension of each head and $N_h$ is the number of heads. Therefore, each head can have different representations, and similarities can be computed per head. For each attention head $h$, a set of linear projections are used to compute the \textit{query} vectors $\mathcal{Q}_h$, \textit{key} vectors $\mathcal{K}_h$ and \textit{value} vectors $\mathcal{V}_h$ using feature descriptors $e_I$ as input:
\begin{equation}
    \begin{split}
        \mathcal{Q}_h &= W_{\mathcal{Q}_h}e_{I} + b_{\mathcal{Q}_h} \\
        \mathcal{K}_h &= W_{\mathcal{K}_h}e_{I} + b_{\mathcal{K}_h} \\
        \mathcal{V}_h &= W_{\mathcal{V}_h}e_{I} + b_{\mathcal{V}_h}\, ,
    \end{split}
\end{equation}
where $W_{\mathcal{Q}_h}, W_{\mathcal{K}_h}, W_{\mathcal{V}_h} \in \mathbb{R}^{C_{h} \times C_{h}}$, $b_{\mathcal{Q}_h}, b_{\mathcal{K}_h}, b_{\mathcal{V}_h} \in \mathbb{R}^{C_{h}}$. We normalize the similarities via softmax and obtain $\alpha_h$ as 
\begin{align}
    \alpha_h &= \text{softmax}(\frac{\mathcal{Q}_h^T \mathcal{K}_h}{\sqrt{C_{h}}}) \,.
    \label{eqn:attn}
\end{align}

\noindent The output \textit{value} vector $\mathcal{V}_\mathcal{O}$ can be computed as:
\begin{align}
    \mathcal{V}_\mathcal{O} = W_\mathcal{O} \, \text{Concat}(\alpha_1 \mathcal{V}_1,...,\alpha_{N_h} \mathcal{V}_{N_h}) + b_\mathcal{O} \, ,
\end{align}
where $W_\mathcal{O} \in \mathbb{R}^{C_{e} \times C_{e}}$ and $b_\mathcal{O} \in \mathbb{R}^{C_{e}}$. The output \textit{value} vector $\mathcal{V}_\mathcal{O}$ is then added to the original feature descriptors to form a residual connection:
\begin{align}
    e_I = e_I + \mathcal{V}_\mathcal{O}\,.
\end{align}

For self-attention, the $\mathcal{Q}_h,\mathcal{K}_h,\mathcal{V}_h$ are computed from the same image. For cross-attention, $\mathcal{Q}_h$ is computed from the source image while $\mathcal{K}_h,\mathcal{V}_h$ are computed from the target image. Cross-attention is applied bi-directionally so in one case source$\rightarrow$target is left$\rightarrow$right and the other case is right$\rightarrow$left.

\subsubsection{Relative Positional Encoding}
\label{sssec:rel_pos_enc}
In large textureless areas, similarities between pixels can be ambiguous. This ambiguity, however, may be resolved by considering relative positional information with respect to prominent features, such as edges. Thus, we provide data-dependent spatial information via positional encoding $e_p$. We choose to encode relative pixel distances instead of absolute pixel locations due to its shift-invariance. In the vanilla Transformer \cite{vaswani2017attention}, the absolute positional encoding $\textcolor{cyan}{e_p}$ is directly added to the feature descriptor 
\begin{equation}
    e = e_I + \textcolor{cyan}{e_p}\,.
\end{equation}
In that case, attention between the $i$-th and $j$-th pixel in \autoref{eqn:attn} can be expanded \cite{dai2019transformer} (ignoring biases for simplicity) as
\begin{multline}
    \alpha_{i,j} = \underbrace{e_{I,i}^TW_\mathcal{Q}^TW_\mathcal{K}e_{I,j}}_\text{(1) data-data}+\underbrace{e_{I,i}^T W_\mathcal{Q}^T W_{K} \textcolor{cyan}{e_{p,j}}}_\text{(2) data-position} + \\  \underbrace{\textcolor{cyan}{e_{p,i}^T}W_\mathcal{Q}^T W_\mathcal{K} e_{I,j}}_\text{(3) position-data} +\underbrace{\textcolor{cyan}{e_{p,i}^T}W_\mathcal{Q}^T W_\mathcal{K} \textcolor{cyan}{e_{p,j}}}_\text{(4) position-position} \,.
\end{multline}
As shown, the term (4) entirely depends on position and should thus be left out, as disparity fundamentally depends on image content. Instead, we use relative positional encoding and remove the term (4) as
\begin{multline}
    \alpha_{i,j} = \underbrace{e_{I,i}^TW_\mathcal{Q}^TW_\mathcal{K}e_{I,j}}_\text{(1) data-data}+ \\ \underbrace{e_{I,i}^T W_\mathcal{Q}^T W_{K} \textcolor{cyan}{e_{p,i-j}}}_\text{(2) data-position} +  \underbrace{\textcolor{cyan}{e_{p,i-j}^T}W_\mathcal{Q}^T W_\mathcal{K} e_{I,j}}_\text{(3) position-data}\,,
    \label{eqn:attn_rel_per_term}
\end{multline}
where $\textcolor{cyan}{e_{p,i-j}}$ denotes the positional encoding between the $i$-th and $j$-th pixel. Note that $\textcolor{cyan}{e_{p,i-j}} \neq \textcolor{cyan}{e_{p,j-i}}$. Intuitively, the attention depends on both content similarity and relative distance. Concurrent to our development, \cite{he2020deberta} found that a similar attention mechanism is beneficial for NLP tasks.

However, the computational cost of relative distance is quadratic in the image width $I_w$ since for each pixel, there are $I_w$ relative distances, and this computation needs to be done $I_w$ times. We describe an efficient implementation which reduces the cost to linear. The details of the implementation follow in \autoref{ap:rel_pos_enc}. 

\subsubsection{Optimal Transport}
\label{sssec:optimal_transport}
Enforcing the uniqueness constraint of stereo matching was attempted in \cite{ohta1985stereo}, where each pixel in the right image gets assigned to at most one pixel in the left image. However, this hard assignment prohibits gradient flow. By contrast, entropy-regularized optimal transport \cite{cuturi2013sinkhorn} is an ideal alternative due to its soft assignment and differentiability, and was previously demonstrated as beneficial for the related task of sparse feature \cite{sarlin2020superglue} and semantic correspondence \cite{liu2020semantic} matching. Given a cost matrix $M$ of two marginal distributions $a$ and $b$ of length $I_w$, the entropy-regularized optimal transport attempts to find the optimal coupling matrix $\mathcal{T}$ by solving
\begin{equation}
    \begin{split}
        \mathcal{T} &= \underset{\mathcal{T}\in R_+^{I_w \times I_w}}  {\text{argmin}}\,\sum_{i,j=1}^{I_w,I_w}\mathcal{T}_{ij}M_{ij} - \gamma E(\mathcal{T}) \\
        \text{s.t.} \ & \mathcal{T}1_{I_w}=a, \ \mathcal{T}^T1_{I_w}=b
    \end{split} \label{eqn:ot}
\end{equation}
where $E(\mathcal{T})$ is the entropy regularization. If the two marginal distributions $a,b$ are uniform, $\mathcal{T}$ is also optimal for the assignment problem, which imposes a soft uniqueness constraint \cite{peyre2019computational} and mitigates ambiguity \cite{liu2020semantic}. The solution to \autoref{eqn:ot} can be found via the iterative Sinkhorn algorithm \cite{cuturi2013sinkhorn}. Intuitively, values in $\mathcal{T}$ represent the pairwise matching probabilities, similar to softmaxed attention in \autoref{eqn:attn}. 
Due to occlusion, some pixels cannot be matched. Following \cite{sarlin2020superglue}, we augment the cost matrix by adding dustbins with a learnable parameter $\phi$ that, intuitively, represents the cost of setting a pixel unmatched.

In STTR, the cost matrix $M$ is set to the negative of the attention computed by the cross-attention module in \autoref{eqn:attn}, but without softmax, as optimal transport will normalize the attention values.
\vspace{-0.1cm}

\subsubsection{Attention Mask}
\label{sssec:attn_mask}
Let $x_L, x_R$ be the projected location of the same physical point onto the left and right epipolar lines respectively ($+x$ from left to the right). The spatial arrangement of the cameras in a stereo rig ensures that $x_R \le x_L$ for all points after rectification. Therefore, in the last cross-attention layer, it is sufficient for each pixel in the left image to only attend to pixels that are further to the left of the same coordinate in the right image (i.e., attend only to points $x$ in the right image where $x \le x_L$). To impose such a constraint, we introduce a lower-triangular binary mask on the attention. Additional visualization can be found in \autoref{ap:attn_mask}.

\subsubsection{Raw Disparity and Occlusion Regression}
In most prior work, a weighted-sum of all candidate disparity values is used. We instead regress disparity using a modified winner-take-all approach \cite{tulyakov2018practical}, which is robust against multi-modal distributions. 

The raw disparity is computed by finding 
the location of the most probable match, denoted as $k$, from the optimal transport assignment matrix $\mathcal{T}$ and build a 3\,px window $\mathcal{N}_{3}(k)$ around it. A re-normalization step is applied to the matching probabilities within the 3\,px window such that the sum is 1. The weighted sum of the candidate disparities is the regressed raw disparity $\tilde{d}_{raw}(k)$. Denoting the matching probability in assignment matrix $\mathcal{T}$ to be $t$, we have

\begin{equation}
    \tilde{t}_l = \frac{t_l}{\sum_{l \in \mathcal{N}_3(k)}t_l} ,\, \text{for} \, l \in \mathcal{N}_{3}(k)
\end{equation}
\begin{equation}
    \tilde{d}_{raw}(k) = \sum_{l \in \mathcal{N}_3(k)} d_l \tilde{t}_l
\end{equation}

The sum of probabilities within this 3\,px window  represents an estimate of the confidence of the network with the current assignment, in the form of an inverse occlusion probability. Therefore, we can regress the occlusion probability $p_{occ}(k)$ using the same information as
\begin{equation}
    p_{occ}(k) = 1- \sum_{l \in \mathcal{N}_3(k)} t_l\,.
\end{equation}

\subsection{Context Adjustment Layer}
\begin{figure}[thpb]
    \centering
    \includegraphics[width=0.9\linewidth, trim={5cm 2cm 5cm 7cm}, clip]{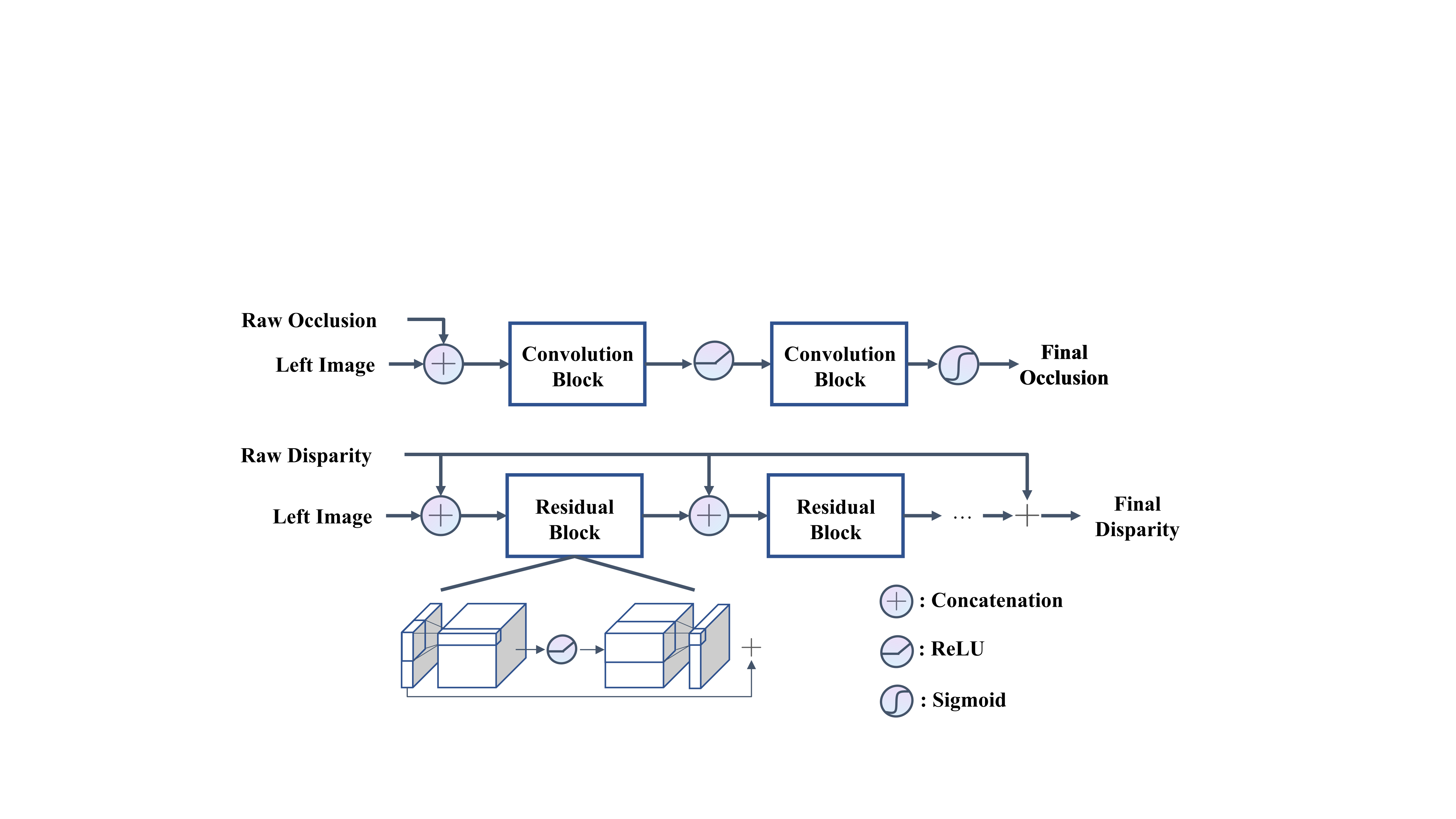}
    \caption{Overview of the context adjustment layer. Convolution blocks and Sigmoid activation are used for occlusion refinement (top) and residual blocks with long skip connections are used for disparity refinement (bottom). Qualitative result is in \autoref{ap:cal}.}
    \label{fig:context_adjustment_layer}
\end{figure}

The raw disparity and occlusion maps are regressed over epipolar lines and thus lack context across multiple epipolar lines. To mitigate this, we use convolutions to adjust the estimated values conditioned on the input image with cross epipolar line information. The overview of the context adjustment layer is in \autoref{fig:context_adjustment_layer}. 

The raw disparity and occlusion maps are first concatenated with the left image along the channel dimension. Two convolution blocks are used to aggregate the occlusion information, followed by ReLU. The final occlusion is estimated by a Sigmoid activation. Disparities are refined by residual blocks which expand the channel dimension before ReLU activation then restores it to the original channel dimension. The expansion before the ReLU is to encourage better information flow \cite{yu2018wide}. Raw disparity is repeatedly concatenated with the residual block for better conditioning. The final output of the residual blocks is added to the raw disparity via a long skip connection. 

\subsection{Loss}
We adopt the Relative Response loss $L_{rr}$ proposed in \cite{liu2020extremely} on the assignment matrix $\mathcal{T}$ for both sets of matched pixels $\mathcal{M}$ and sets of unmatched pixels $\mathcal{U}$ due to occlusion. The goal of the network is to maximize the attention on the true target location. Since disparity is subpixel, we use linear interpolation between the nearest integer pixels to find the matching probability $t^*$. Specifically, for the $i$-th pixel in the left image with ground truth disparity $d_{gt,i}$,
\begin{equation}
    \begin{split}
        t_i^* &= \text{interp}(\mathcal{T}_i, p_i - d_{gt,i}) \\
        L_{rr} &= \frac{1}{N_{\mathcal{M}}}\sum_{i \in \mathcal{M}}-log(t_i^*) + \frac{1}{N_{\mathcal{U}}}\sum_{i \in \mathcal{U}}-log(t_{i,\phi})
    \end{split}
\end{equation}
where \textit{interp} denotes linear interpolation and $t_{i,\phi}$ is the unmatched probability. We use smooth L1 loss \cite{girshick2015fast} on both raw and final disparities, denoted as $L_{d1,r}$ and $L_{d1,f}$. The final occlusion map is supervised via a binary-entropy loss $L_{be, f}$. The total loss is the summation:
\begin{equation}
    L = w_1L_{rr} + w_2L_{d1, r} + w_3L_{d1,f} + w_4L_{be, f}\, ,
\end{equation}
where $w$ are the loss weights.

\subsection{Memory-Feasible Implementation}
\label{ssec:mem_eff}
The memory consumption of the attention mechanism is quadratic in terms of the sequence length. Specifically, for a float32 precision computation,
\begin{equation}
    \text{memory consumption in bits} = 32I_hI^2_wN_{h}N \,.
\end{equation}
For example, given $I_w=960$, $I_h=540$ and $N_h=8$, training a $N=6$ layer Transformer consume approximately $216$~GB, which is impractical on conventional hardware. Following \cite{child2019generating}, we adopt gradient checkpointing \cite{griewank2000algorithm} for each self- and cross-attention layer, where the intermediate variables are not saved during the forward pass. During the backward pass, we run the forward pass again for the checkpointed layer to recompute the gradient. Thus, the memory consumption is bounded by the requirement of a single attention layer, which in theory enables the network to scale infinitely in terms of the number of attention layers $N$.

Moreover, we use mixed-precision training \cite{micikevicius2017mixed} for faster training speed and reduced memory consumption.

Lastly, we use an attention stride $s>1$ to sparsely sample the feature descriptors, which is equivalent to a downsampling of the feature map.

\textbf{Complexity Analysis:} In existing cost-volume paradigms, correlation-based networks have a memory complexity of $\mathcal{O}(I_h I_w D)$, while 3D convolution-based networks have $\mathcal{O}(I_h I_w D C)$, where $D$ is the maximum disparity value and $C$ is the channel size. $D$ is generally set to a fixed value less than $I_w$, sacrificing the ability to predict disparity values outside the range. STTR is of $\mathcal{O}(I_h I_w^2 / s^3)$, which offers an alternative trade-off where no maximum disparity is set. Given $s$, STTR runs at a \textit{constant} memory consumption across \textit{different} disparity ranges compared to prior work. During inference, $s$ can be adjusted to a larger value which reduces memory consumption and maintains the maximal disparity range at the slight sacrifice of task performance. Quantitative analysis of the trade-off between performance and memory of $s$ is in \autoref{ap:attn_stride}. We also introduce a lightweight implementation of STTR without the flexibility to adjust $s$ in \autoref{ap:lightweight} for faster speed and lower memory consumption. Comparison of inference speed/memory between STTR and prior work is in \autoref{ap:inference_stats}.

\section{Experiments, Results, and Discussion}
\label{sec:exp}
\textbf{Datasets:} Scene Flow \cite{mayer2016large} FlyingThings3D subset is a synthetic dataset of random objects. MPI Sintel \cite{butler2012naturalistic} is a synthetic dataset from animated film that contains various realistic artifacts such as specular reflections and motion blur. KITTI 2015 \cite{menze2015object} is a street scene dataset. Middlebury 2014 \cite{scharstein2014high} quarter resolution subset is an indoor scene dataset. SCARED \cite{allan2021stereo} is a medical scene dataset of laparoscopic surgery. For pre-training, we use the default split of Scene Flow. For cross-domain generalization evaluation, we use all the provided data from each dataset. For KITTI 2015 benchmark evaluation, we train on KITTI 2012 and 2015 dataset and leave 20 images for validation. Details of the datasets and pre-processing step are in \autoref{ap:dataset_info}. Training duration and number of parameters are in \autoref{ap:training_param}.

\textbf{Hyperparameters:} In our experiments, we use 6 self- and cross-attention layers with $C_e=128$. We run the Sinkhorn algorithm for 10 iterations. We use attention stride of $s=3$ during training. We use AdamW as the optimizer with weight decay of 1e-4. We set all loss weights $w$ to 1. We pre-train on Scene Flow for 15 epochs using a fixed learning rate of 1e-4 for feature extractor and Transformer, and 2e-4 for the context adjustment layer. To simulate realistic stereo artifacts, we use asymmetric (i.\,e. different for left and right images) augmentation, including RGB shift, Gaussian noise, brightness/contrast shift, vertical shift and rotation. For KITTI 2015 benchmark submission, we fine-tune the pre-trained model using the exponential learning rate scheduler with a decay of 0.99 for 400 epochs. We conduct our experiments on one Nvidia Titan RTX GPU. We use both 3\,px Error (percentage of errors larger than 3\,px) and EPE (absolute error) as evaluation metrics. Please note that all quantitative metrics reported in the remainder of this section refer to \textit{non-occluded regions} only; we use IOU to evaluate \textit{occlusion estimation}.

\subsection{Ablation Studies}
We conduct ablation studies using the Scene Flow synthetic dataset, and provide quantitative results for the effects of the attention mask, optimal transport layer, context adjustment layer and positional encoding summarized in \autoref{tab:ablation}. Following prior work, we validate on the test split directly since Scene Flow is only used for pre-training.

\begin{table}[htpb]
    \centering
    \caption{Ablation study on Scene Flow dataset. AM: attention mask. OT: optimal transport. CAL: context adjustment layer. RPE: relative positional encoding.}
    \resizebox{0.7\linewidth}{!}{%
        \begin{tabular}{cccc||c|c|c}
            \multicolumn{4}{c||}{Component}& 3\,px &  & Occ\\ \cline{1-4}
            AM & OT & CAL & RPE & Error $\downarrow$ & EPE $\downarrow$ & IOU $\uparrow$ \\ \hline
            & &  &  & 3.61 & 0.92  & 0.78\\  \hline
            \checkmark & & & & 2.77 & 0.84 & 0.77 \\ \hline
            \checkmark & \checkmark & &  & 2.32 & 0.70 & 0.87 \\ \hline
            \checkmark & \checkmark & \checkmark &  & 2.21 & 0.63 & 0.88 \\ \hline
            \checkmark & \checkmark & \checkmark  & \checkmark & \textbf{1.26} & \textbf{0.45} & \textbf{0.92}
        \end{tabular}
    }
    \label{tab:ablation}
\end{table}

\textbf{Uniqueness Constraint:} The soft uniqueness constraint is imposed via the optimal transport layer by taking the interaction between pixels along the same epipolar line into account. We find it improves the result in all metrics, especially for occlusion IOU (row 3 in \autoref{tab:ablation}). 

\begin{figure}[htpb]
    \centering
    \subfloat[Left image.]{\includegraphics[height=0.18\linewidth,trim={-2cm 0cm -2cm 0cm},clip]{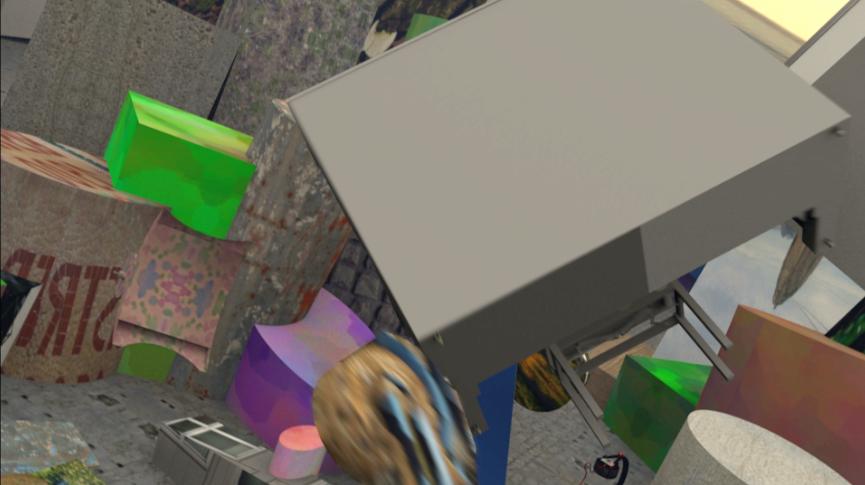}}
    \hspace{0.5em}
    \subfloat[Output of feature extractor.]{\includegraphics[height=0.18\linewidth,trim={-3.5cm 2cm -2.5cm 2cm},clip]{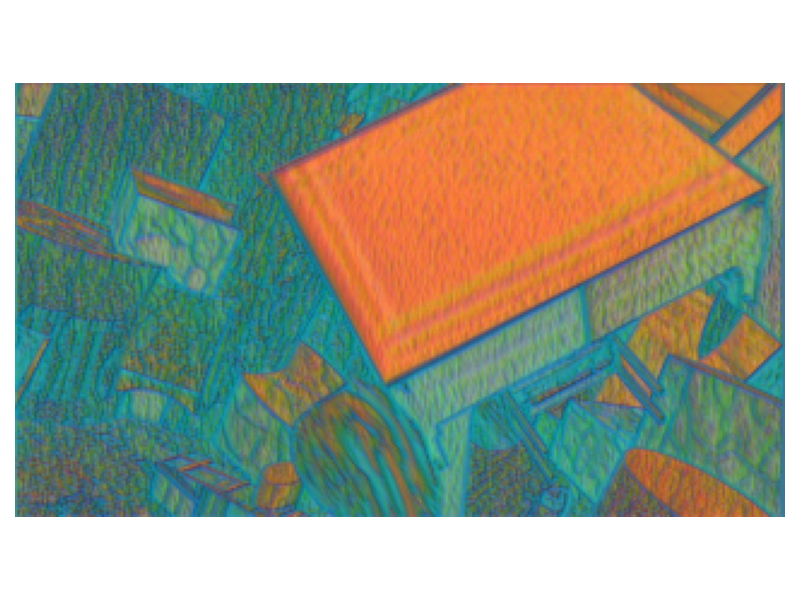}}
    
    \subfloat[Output of Transformer layer 4, \textbf{without} position encoding.]{\includegraphics[height=0.18\linewidth,trim={-3cm 2cm -3cm 2cm},clip]{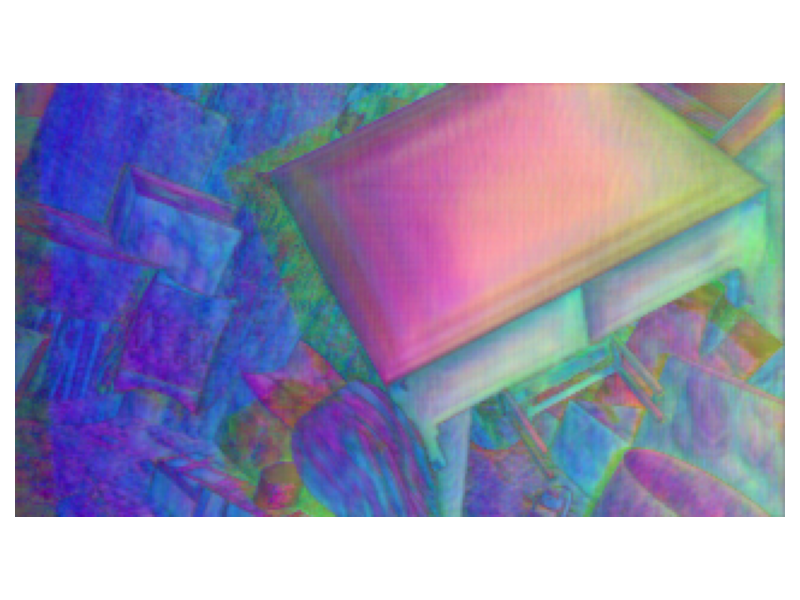}}
    \hspace{0.5em}
    \subfloat[Output of Transformer layer 6, \textbf{without} position encoding.]{\includegraphics[height=0.18\linewidth,trim={-3cm 2cm -3.5cm 2cm},clip]{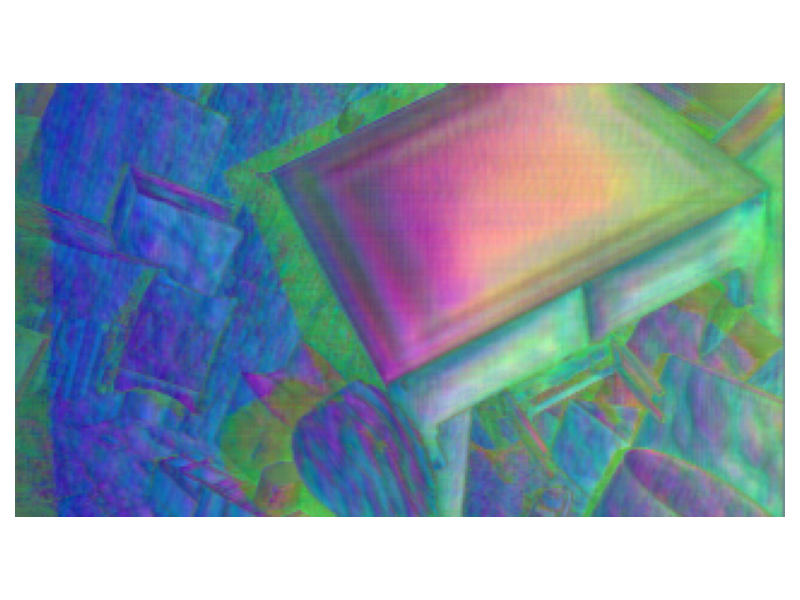}}
    
    \subfloat[Output of Transformer layer 4, \textbf{with} position encoding.]{\includegraphics[height=0.18\linewidth,trim={-3cm 2cm -3.5cm 2cm},clip]{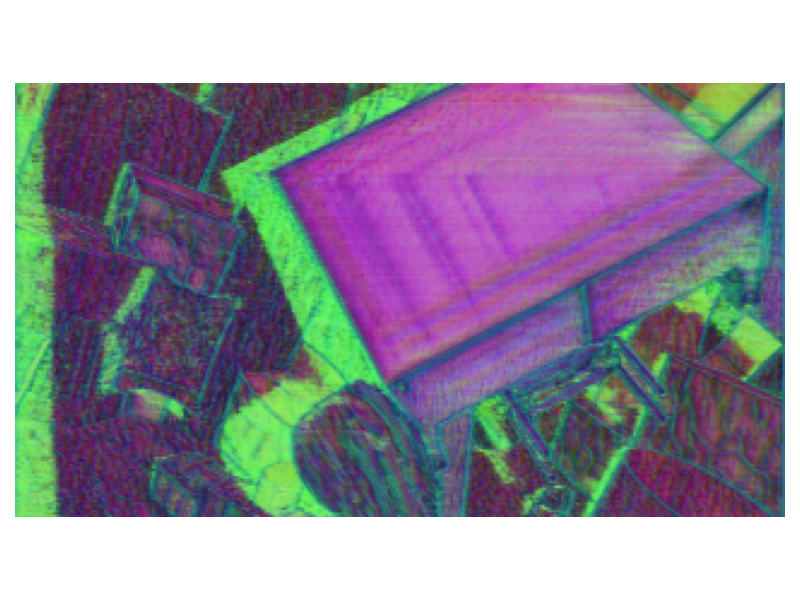}}
    \hspace{0.5em}
    \subfloat[Output of Transformer layer 6, \textbf{with} position encoding.]{\includegraphics[height=0.18\linewidth,trim={-3cm 2cm -3cm 2cm},clip]{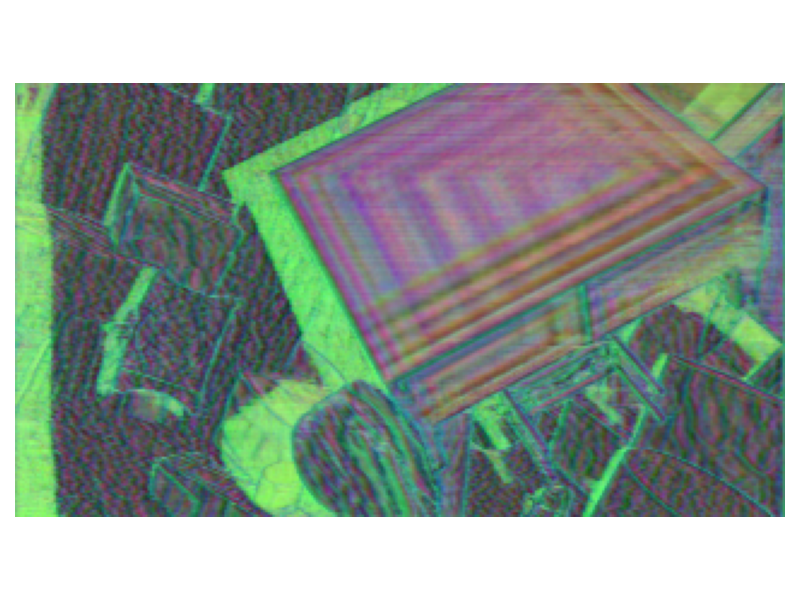}}
    
    \caption{Feature descriptor visualization. Full evolution of features updated by Transformer can be found in \autoref{ap:evolution_of_feature_map}.}
    \label{fig:feat_vis}
\end{figure}

\textbf{Relative Positional Encoding:}
To visualize the effect of positional encoding, we use PCA to reduce the feature map to $\mathbb{R}^3$. Given an image with a large textureless area, such as the table shown in \autoref{fig:feat_vis}(a), the features directly extracted from the feature extractor shown in \autoref{fig:feat_vis}(b) exhibit similar patterns. In the case of no positional encoding, as the layers progress deeper, the feature map merely changes throughout the process as shown in \autoref{fig:feat_vis}(c-d). By providing relative positional encoding to all layers, strides that are parallel to the edges emerge in layer 4 as shown in \autoref{fig:feat_vis}(e) and eventually the strides propagate to the entire region in layer 6 as shown in \autoref{fig:feat_vis}(f). This suggests that the Transformer needs relative positional information to resolve ambiguity in textureless areas. With positional encoding added to all layers, the result improves for all three metrics (row 5 in \autoref{tab:ablation}).

\textbf{Generalization of Attention:}
In principle, STTR allows the disparity range to scale with the image width since pixels are densely compared. Nonetheless, we evaluate if STTR indeed generalizes beyond the disparity range that it is trained on. We train another model by only computing losses on pixels with disparities smaller than 192 px (0.2$I_w$) and ignoring pixels outside this range. During testing, the maximal disparity prediction made within 1\,px error margin is 458\,px (0.48$I_w$), which demonstrates the generalization.

\textbf{Attention Span:}
We analyze the attention span (i.\,e., the distribution of attention values over all pixels) of each layer of self- and cross-attention \cite{sukhbaatar2019adaptive,tay2020long}. Our results (detailed in \autoref{ap:evolution_of_attention}) show that both self- and cross-attention start from relatively global context (300 px, 0.31$I_w$) and shift to the local context (114 px, 0.12$I_w$ for self-attention and 15 px, 0.01$I_w$ for cross-attention). Since the attention span decreases in deeper layers, we conclude that the global context is predominantly used in early layers but does not contribute substantially to disparity refinement in the late layers. This suggests an opportunity for future work to gradually reduce the search window based on previous layers' attention span to improve efficiency. 

\begin{table*}[htpb]
    \centering
    \caption{Generalization \textit{without} fine-tuning on MPI Sintel, KITTI 2015, Middlebury 2014, and SCARED dataset. \textbf{Bold} is best. $\ddagger$: models trained with asymmetric data augmentation. $\dagger$: $s=4$ for STTR due to memory constraint. OOM: out-of-memory. (W$\times$H): image resolution.}
    \resizebox{0.95\linewidth}{!}{%
    \begin{tabular}{c||c|c|c||c|c|c||c|c|c||c|c|c}
     & \multicolumn{3}{c||}{MPI Sintel~$\dagger$ ($1024\times436$)} & \multicolumn{3}{c||}{KITTI 2015 ($1242\times375$)} & \multicolumn{3}{c||}{Middlebury 2014 (varies)} & \multicolumn{3}{c}{SCARED~$\dagger$ ($1080\times1024$)} \\ \cline{2-13} 
     & 3\,px Error $\downarrow$ & EPE $\downarrow$ & Occ IOU $\uparrow$ & 3\,px Error $\downarrow$ & EPE $\downarrow$ & Occ IOU $\uparrow$ & 3\,px Error $\downarrow$ & EPE $\downarrow$ & Occ IOU $\uparrow$ & 3\,px Error $\downarrow$ & EPE $\downarrow$ & Occ IOU $\uparrow$ \\ \hline 
    PSMNet \cite{chang2018pyramid} & 6.81 & 3.31 & N/A & 27.79 & 6.56 & N/A & 12.96 & 3.05 & N/A & OOM & OOM & N/A \\
    PSMNet~$\ddagger$ & 7.93 & 3.70 & N/A & 7.43 & 1.39 & N/A & 10.24 & 2.02 & N/A & OOM & OOM & N/A \\ 
    GwcNet-g \cite{guo2019group} & 6.26 & 1.42 & N/A & 12.60
	 & 2.21 & N/A & 8.59 & 1.89 & N/A & OOM & OOM & N/A \\
    GwcNet-g~$\ddagger$ & 5.83 & \textbf{1.32} & N/A & 6.75 & 1.59 & N/A & 6.60 & 1.95 & N/A & OOM & OOM & N/A \\
    AANet \cite{xu2020aanet} & 5.91 & 1.89 & N/A & 12.42 & 1.99 & N/A & 12.80 & 2.19 & N/A & 6.39 & 1.36 & N/A \\
    AANet~$\ddagger$ & 6.29 & 2.24 & N/A & 7.06 & \textbf{1.31} & N/A & 9.57 & \textbf{1.71} & N/A & 3.99 & \textbf{1.17} & N/A \\ \hline
    \textbf{STTR~$\ddagger$} & \textbf{5.75} & 3.01 & \textbf{0.86} & \textbf{6.74} & 1.50 & \textbf{0.98} & \textbf{6.19} & 2.33 & \textbf{0.95} & \textbf{3.69} & 1.57 & \textbf{0.96}
    \end{tabular}
    }%
    \label{tab:generalization}
\end{table*}

\subsection{Comparison with Prior Work}
Under multiple evaluation settings, we compare STTR with prior work spanning the primary learning-based stereo depth paradigms, including correlation-based AANet \cite{xu2020aanet}, 3D convolution-based PSMNet \cite{chang2018pyramid} and GANet-11 \cite{zhang2019ga}, correlation and 3D convolution hybrid approach GwcNet-g \cite{guo2019group} and a classification-based Bi3D \cite{badki2020bi3d}.
\vspace{-0.2cm}

\subsubsection{Scene Flow Benchmark Result}
The pre-training results on Scene Flow are shown in \autoref{tab:sceneflow}. STTR performs on par with prior work when evaluated only on pixels with disparity less than 192 (2nd -- 3rd columns). However, STTR outperforms prior work by a large margin in unconstrained settings due to its unbounded disparity estimation (4th -- 5th columns) while the maximal disparity is fixed at $D=192$ for the other methods. To compare fairly, we evaluate again with $D=480$ (which covers all disparity values in the test dataset) for prior work. STTR's performance remains the \textit{same} as $D=192$ setting and comparable to prior work. Moreover, as shown qualitatively in \autoref{fig:overview} and quantitatively in \autoref{tab:ablation}, STTR can accurately identify occluded areas, which was not attempted in prior work. 
\vspace{-0.2cm}

\begin{table}[htpb]
\centering
\caption{Evaluation on Scene Flow. Model weights provided by authors. D: maximum disparity value. OOM: out of memory. N/A: model only infers within 192 range.}
\label{tab:sceneflow}
\resizebox{\linewidth}{!}{%
\begin{tabular}{c||cc|cc||cc|cc}
 & \multicolumn{4}{c||}{D=192} & \multicolumn{4}{c}{D=480} \\ \cline{2-9} 
 & \multicolumn{2}{c|}{disp \textless 192} & \multicolumn{2}{c||}{All pixels} & \multicolumn{2}{c|}{disp \textless 192} & \multicolumn{2}{c}{All pixels} \\ \cline{2-9} 
 & 3 px  &  & 3 px & & 3 px  & & 3 px &  \\
 & Error $\downarrow$ & EPE $\downarrow$ & Error $\downarrow$ & EPE $\downarrow$ & Error $\downarrow$ & EPE $\downarrow$ & Error $\downarrow$ & EPE $\downarrow$ \\ \hline
PSMNet & 2.87 & 0.95 & 3.31 & 1.25 & 3.09 & 0.92 & 3.60 & 1.03 \\
GwcNet-g & 1.57 & 0.48 & 2.09 & 0.89 & 1.60 & 0.50 & 1.72 & 0.53 \\
AANet & 1.86 & 0.49 & 2.38 & 1.96 & \multicolumn{2}{c|}{N/A} & \multicolumn{2}{c}{N/A} \\
GANet-11 & 1.60 & 0.48 & 2.19 & 0.97 & \multicolumn{2}{c|}{OOM} & \multicolumn{2}{c}{OOM} \\
Bi3D & 1.70 & 0.54 & 2.21 & 1.16 & \multicolumn{2}{c|}{OOM} & \multicolumn{2}{c}{OOM} \\ \hline
\textbf{STTR} & \textbf{1.13} & \textbf{0.42} & \textbf{1.26} & \textbf{0.45} & \textbf{1.13} & \textbf{0.42} & \textbf{1.26} & \textbf{0.45}
\end{tabular}%
}
\end{table}

\subsubsection{Cross-Domain Generalization}
We examine the domain generalization of STTR by comparing it to prior work also trained only on the Scene Flow synthetic dataset in \autoref{tab:generalization}. Note that we do not refine the models to the test dataset. For a fair comparison, we set maximum disparity $D=192$ for prior work and only evaluate pixels in this range. We also trained prior work with the same asymmetric augmentation technique used for STTR to avoid inconsistency \cite{watson2020learning}, while keeping the original training scheme (optimizer, learning rate, loss and number of epochs). Although asymmetric augmentation improves generalization in some cases (confirming findings of \cite{watson2020learning}), it is not consistent. Regardless, STTR generalizes comparably and maintains a high occlusion IOU across four datasets. STTR performs well in terms of 3\,px Error, but not always for EPE. This is because if a pixel is wrongly identified as occlusion, a zero disparity is predicted, resulting in a large EPE. We visualize the generalization mechanism of STTR in \autoref{ap:generalization}.

\subsubsection{KITTI Benchmark Result}
Since the KITTI benchmark provides a reasonable number of images for fine-tuning and it is commonly used in prior work, we choose the KITTI benchmark for comparisons after fine-tuning. The result on KITTI 2015 benchmark is shown in \autoref{tab:benchmark}\footnote{\href{http://www.cvlibs.net/datasets/kitti/eval_scene_flow_detail.php?benchmark=stereo&result=c702741f545d79c4ba985800dfc2c87a8f8af1a0}{Full result of KITTI benchmark.}}. STTR performs comparably to several competing approaches, even compared with multi-resolution networks designed for better context aggregation. 

\subsubsection{Shortcomings in Challenge Design}
While STTR performs comparatively to prior work, it is worth mentioning that the test sets in these real-world datasets are rather small. Specifically, the test sets accompanying KITTI 2015, Middlebury 2014, and SCARED contain 200, 15, and 19 images, respectively. To achieve a more comprehensive estimate of performance, much larger test datasets are required. This paucity of data is further compounded by the observation that performance differences between competing models on these datasets are on the order of $<1\,\%$ after refinement and a few percent for cross-domain generalization. Consequently, we cannot conclude if there is a significant performance difference between the approaches presented. Additionally, KITTI only reports on disparities less than 192 and does not have metrics related to some of the benefits central to STTR (i.\,e., unlimited disparity and occlusion detection). As such, results on this benchmark likely cannot give a complete picture of performance comparisons. 

Despite the above limitation, certain advancements have brought substantial performance improvements for cost-volume approaches. The shift from low-res (e.\,g. PSMNet \cite{chang2018pyramid}) to multi-res using feature pyramids for better context aggregation (such as LEAStereo \cite{cheng2020hierarchical}) appears to be a fruitful path, where LEAStereo \cite{cheng2020hierarchical} is further optimized within this structure using neural architecture search. Our future work aims to incorporate these developments into our design.

\begin{table}[htpb]
    \centering
    \caption{Evaluation of 3\,px or 5$\%$ Error on KITTI 2015. bg: background. fg: foreground. multi-res: networks operate on multiple resolutions. low-res: networks operate on downsampled resolution.}
    \resizebox{0.7\linewidth}{!}{%
        \begin{tabular}{c||c|c||c|c|c}
                    & Methods & Year & bg $\downarrow$ & fg $\downarrow$ & all $\downarrow$ \\ \hline
            \multirow{4}{*}{multi-res} 
            & HSM-1.8 \cite{yang2019hierarchical} & 2019 & 1.63 & 3.40 & 1.92 \\
            & AMNet \cite{du2019amnet} & 2019 & 1.39 & 3.20 & 1.69\\
            & AANet \cite{xu2020aanet} & 2020 & 1.80 & 4.93 & 2.32 \\
            & LEAStereo \cite{cheng2020hierarchical} & 2020 & \textbf{1.29} & \textbf{2.65} & \textbf{1.51} \\
            \hline
            \multirow{5}{*}{low-res} & PSMNet \cite{chang2018pyramid} & 2018 & 1.71 & 4.31 & 2.14 \\
            & GANet-15 \cite{zhang2019ga} & 2019 & 1.40 & 3.37 & 1.73 \\
            & GwcNet-g \cite{guo2019group} & 2019 & 1.61 & 3.49 & 1.92\\
            & Bi3D \cite{badki2020bi3d} & 2020 & 1.79 & 3.11 &2.01 \\ 
            & \multicolumn{2}{c||}{\textbf{STTR}} & 1.70 & 3.61 & 2.01  
        \end{tabular}
    }
    \label{tab:benchmark}
\end{table}

\section{Conclusion}
In conclusion, we have presented an end-to-end network architecture named STereo TRansformer that synergizes advantages of CNN and Transformer architectures. We revisit stereo depth estimation from a sequence-to-sequence matching perspective. This approach 1) avoids the need to pre-specify a fixed disparity range, 2) explicitly handles occlusion, and 3) imposes a match uniqueness constraint. We experimentally demonstrate that STTR generalizes to different domains without fine-tuning and report promising results on benchmarks with refinement. Future work will include increasing the context information via multi-resolution techniques. 

\appendix
\textbf{Acknowledgements.} We thank anonymous reviewers for their constructive comments. This work was funded in part by a research contract from Galen Robotics and in part by Johns Hopkins University internal funds.

\textbf{Disclosure.} Under a license agreement between Galen Robotics, Inc. and the Johns Hopkins University, Dr. Russell H. Taylor and the University are entitled to royalty distributions on technology related to technology described in the study discussed in this publication. Dr. Russell H. Taylor also is a paid consultant to and owns equity in Galen Robotics, Inc. This arrangement has been reviewed and approved by the Johns Hopkins University in accordance with its conflict-of-interest policies.

\section*{Appendix}
\section{Efficient Implementation of Attention with Relative Positional Encoding}
\label{ap:rel_pos_enc}
The attention with relative positional encoding is computed as 
\begin{multline*}
    \alpha_{i,j} = \underbrace{e_{I,i}^TW_\mathcal{Q}^TW_\mathcal{K}e_{I,j}}_\text{(1) data-data}+ \\ \underbrace{e_{I,i}^T W_\mathcal{Q}^T W_{K} \textcolor{cyan}{e_{p,i-j}}}_\text{(2) data-position} +  \underbrace{\textcolor{cyan}{e_{p,i-j}^T}W_\mathcal{Q}^T W_\mathcal{K} e_{I,j}}_\text{(3) position-data}\,.
\end{multline*}

In the following context, we use term (2) data-position (D2P) as an example. For simplicity, we denote $e_{I,i}^TW_\mathcal{Q}^T$ as $\mathcal{Q}_i$ and $W_\mathcal{K}\textcolor{cyan}{e_{p,i-j}}$ as $\textcolor{cyan}{K_{r,i-j}}$. In term (2) for any $i,j$, the relative distance follows a fixed pattern \cite{dai2019transformer} shown as
\begin{equation*}
    \text{D2P} = \left[
            \begin{smallmatrix}
                \mathcal{Q}_0 \textcolor{cyan}{K_{r,0}} & \mathcal{Q}_0 \textcolor{cyan}{K_{r,-1}} & \cdots
                & \mathcal{Q}_0 \textcolor{cyan}{K_{r,-I_w+1}}\\
                \mathcal{Q}_1 \textcolor{cyan}{K_{r,1}} & \mathcal{Q}_1 \textcolor{cyan}{K_{r,0}} & \cdots
                & \mathcal{Q}_1 \textcolor{cyan}{K_{r,-I_w+2}}\\
                & &\vdots & \\
                \mathcal{Q}_{I_w-1} \textcolor{cyan}{K_{r,I_w-1}} & \mathcal{Q}_{I_w-1} \textcolor{cyan}{K_{r,I_w-2}} & \cdots
                & \mathcal{Q}_{I_w-1} \textcolor{cyan}{K_{r,0}}
            \end{smallmatrix}
        \right]
\end{equation*}
where the first row starts with a relative distance of 0 and ends with a relative distance of $-I_w+1$. Subsequent rows simply offset the first row by increasing amounts. Furthermore, since the relative distance is bounded by $I_w-1$ and $-I_w+1$, therefore we can pre-compute all possible values of $\textcolor{cyan}{K_{r,i-j}}$ as
\begin{equation*}
    \tilde{K}_r = \left[
            \begin{smallmatrix}
                \textcolor{cyan}{K_{r,I_w-1}} \\
                \textcolor{cyan}{K_{r,I_w-2}} \\
                \vdots \\
                \textcolor{cyan}{K_{r,-I_w+1}}
            \end{smallmatrix}
        \right]
\end{equation*}
We then slice $\tilde{K}_r$ with an increasing offset, which can be done computation-lightly to obtain
\begin{equation*}
    K_r = \left[
            \begin{smallmatrix}
                \textcolor{cyan}{K_{r,0}} & \textcolor{cyan}{K_{r,-1}} & \cdots
                & \textcolor{cyan}{K_{r,-I_w+1}}\\
                \textcolor{cyan}{K_{r,1}} & \textcolor{cyan}{K_{r,0}} & \cdots
                & \textcolor{cyan}{K_{r,-I_w+2}}\\
                & &\vdots & \\
                 \textcolor{cyan}{K_{r,I_w-1}} & \textcolor{cyan}{K_{r,I_w-2}} & \cdots
                & \textcolor{cyan}{K_{r,0}}
            \end{smallmatrix}
        \right]
\end{equation*}
The term (2) D2P can then be computed as a matrix product between $\mathcal{Q}$ and $K_r$. The term (3) position-data can be computed similarly, thus, reducing the computation cost of relative position.

\section{Attention Mask}
\label{ap:attn_mask}
\begin{figure}
    \centering
    \subfloat[Left Right Geometry Visualization]{\includegraphics[width=0.65\linewidth,trim={8cm 4cm 8cm 2.5cm},clip]{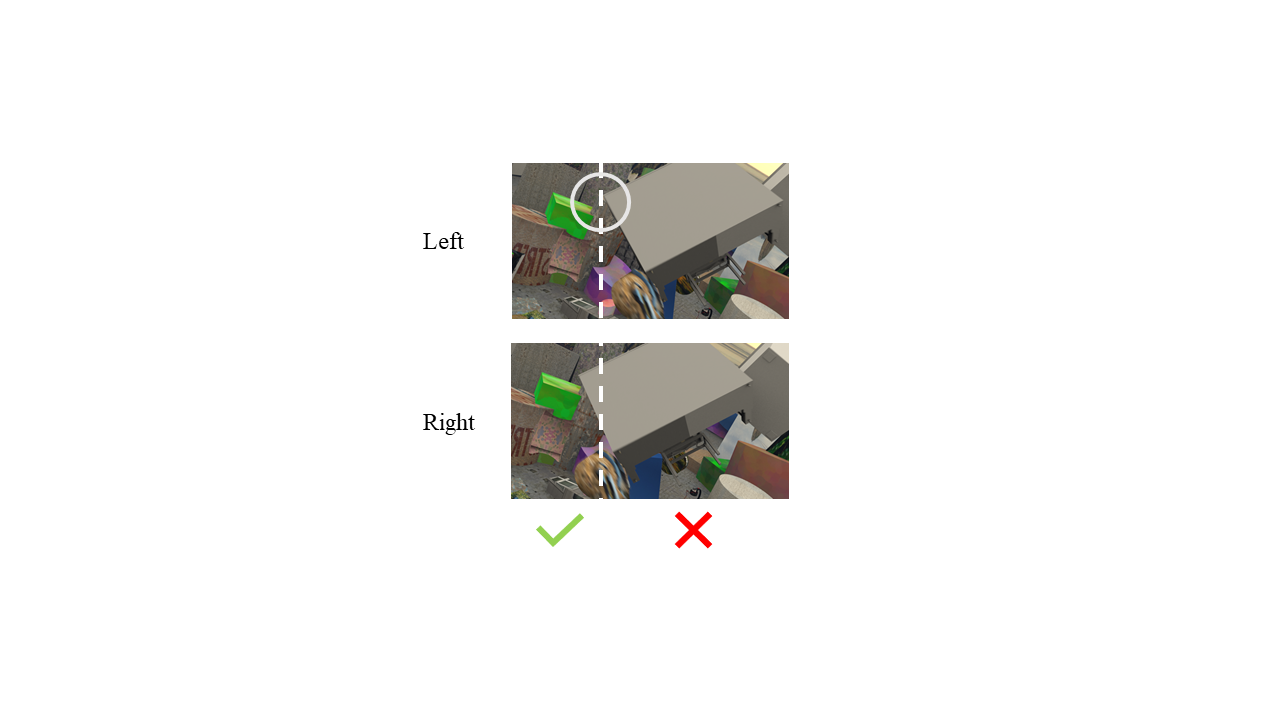}}
    \subfloat[Attention Mask]{\includegraphics[width=0.3\linewidth,trim={1cm 1.5cm 1cm 1.5cm},clip]{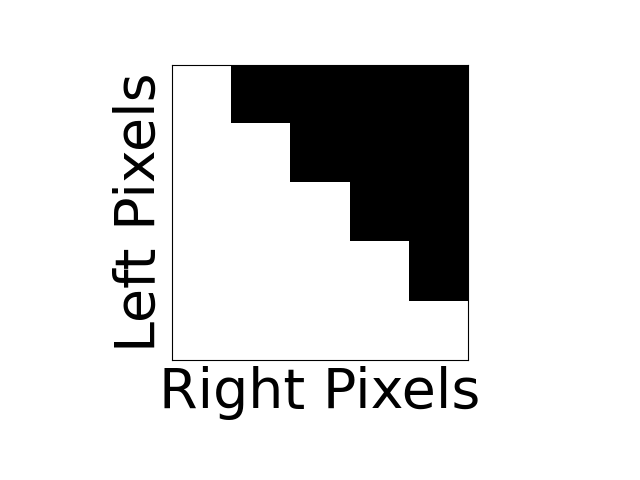}}
    \caption{Attention mask visualization, where white indicates allowable attention region while black indicates forbidden attention region.}
    \label{fig:attn_mask}
\end{figure}

As described in Sect.~3.2.4, if $x_L, x_R$ represent the x-coordinates in pixel-space for the projection of a physical point into the left and right images, then $x_R \leq x_L$ for all physical points (with positive x-axis pointing to the right). Therefore when searching for correspondences, the network only needs to search for pixels in the right image that are to the left of the current pixel position in the left image. An example is given in \autoref{fig:attn_mask}(a): If the corner of the table (highlighted with the circle) is to be matched, the network only needs to search to the left of the dashed line in the right image. This can be achieved using a binary attention mask shown in \autoref{fig:attn_mask}(b), where for each pixel in the left image, the allowable attended pixel locations in the right image marked as white are to the left of source pixel location. Mathematically, this binary attention mask can be achieved by setting the values of forbidden attended pixels (marked as black in \autoref{fig:attn_mask}(b)) in the attention matrix to negative infinity. Thus, after softmax, the attention values $\alpha_h$ in Equation 2 of those pixels will be zero and they'll be excluded from the disparity computation.

\section{Relative Positional Encoding}
\label{ap:evolution_of_feature_map}
\begin{figure*}[htpb]
    \centering
    \includegraphics[width=0.9\linewidth,trim={2.5cm 6.5cm 3cm 4.5cm},clip]{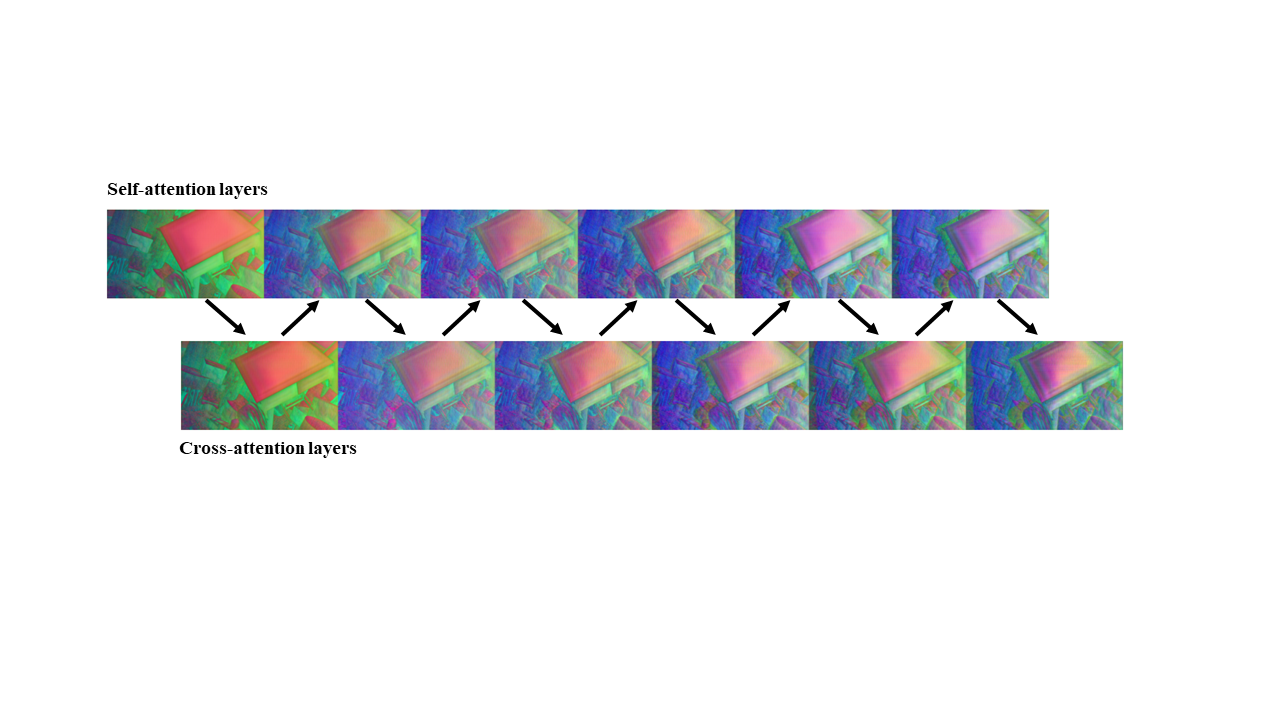}
    
    \caption{Evolution of feature map \textbf{without positional encoding}. First row - input to Transformer self-attention layers 1-6. Second row - input to Transformer cross-attention layers 1-6.}
    \label{fig:feat_ev_no_pos}
\end{figure*}

In \autoref{fig:feat_ev_no_pos}, we visualize the evolution of feature descriptors as the Transformer updates them \textbf{without the positional encoding}. It is worth noting the textureless region, such as the table, does not have distinct patterns to resolve matching ambiguities.

\begin{figure*}[htpb]
    \centering
    \includegraphics[width=0.9\linewidth,trim={2.5cm 6.5cm 3cm 4.5cm},clip]{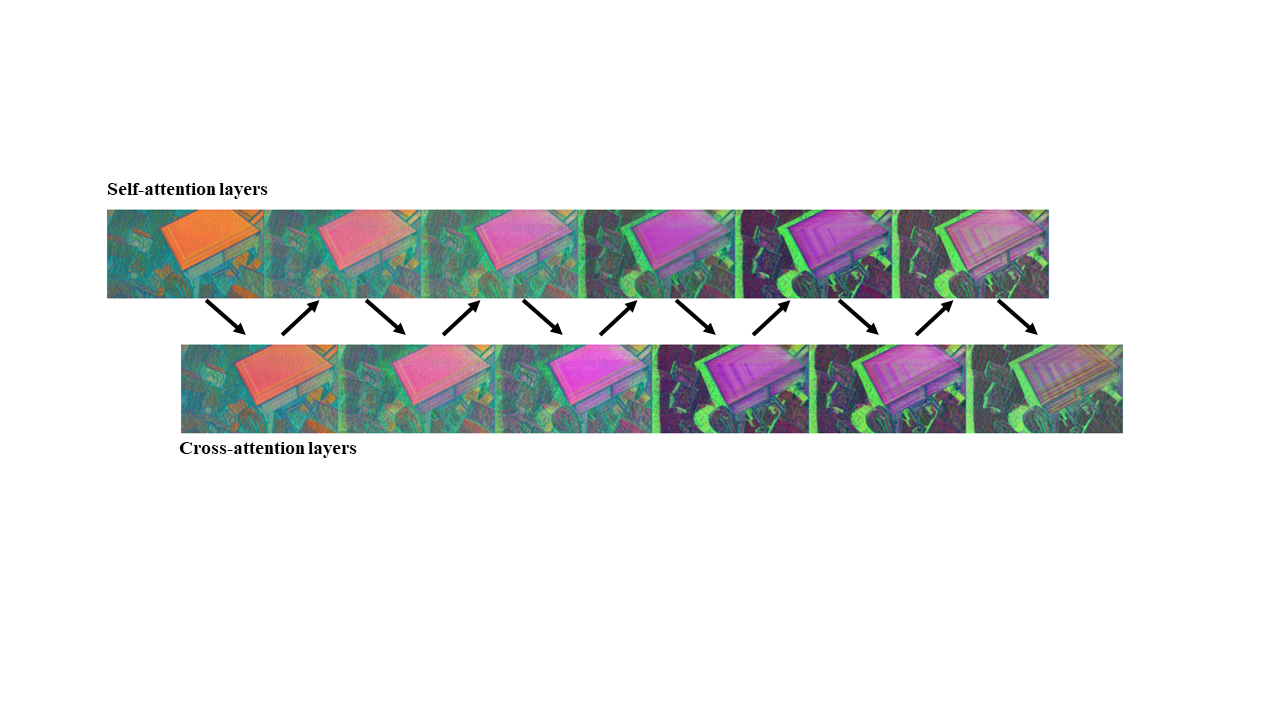}
    
    \caption{Evolution of feature map \textbf{with positional encoding}. First row - input to Transformer self-attention layers 1-6. Second row - input to Transformer cross-attention layers 1-6.}
    \label{fig:feat_ev_pos}
\end{figure*}

In contrast, the evolution of feature descriptors \textbf{with positional encoding} in \autoref{fig:feat_ev_pos} propagates the edge information into the center of the textureless region progressively, which helps to resolve the ambiguity.

\section{Attention Span}
\label{ap:evolution_of_attention}
\begin{figure}[h]
    \centering
    \includegraphics[width=0.55\linewidth]{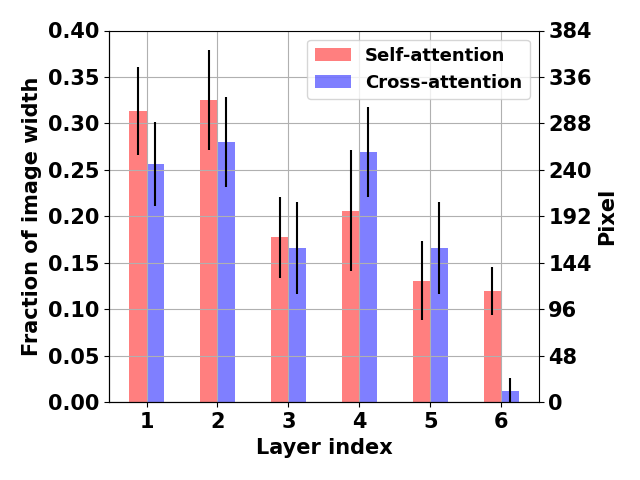}
    \caption{Attention span of both self- and cross-attention evaluated on Scene Flow dataset. Image resolution 960$\times$540. Left: attention span in fraction of image width. Right: attention span in pixels.}
    \label{fig:attn_span}
\end{figure}

We compute attention span of both the self- and cross-attention layers, which is the spatial span of pixels that are above the uniform attention value (i.\,e., 1/$I_w$ with $I_w$ being image width). The layer-wise attention span in~\autoref{fig:attn_span} illustrates how self- and cross-attention shift from a global context to local one as processing moves to higher layers in the network. This is particularly true for cross-attention, where the final attention span is only around 15 pixels (0.01 of image width).

\begin{figure*}[htpb]
    \centering
    \subfloat{\includegraphics[width=0.16\linewidth,trim={2cm 2cm 2cm 2cm},clip]{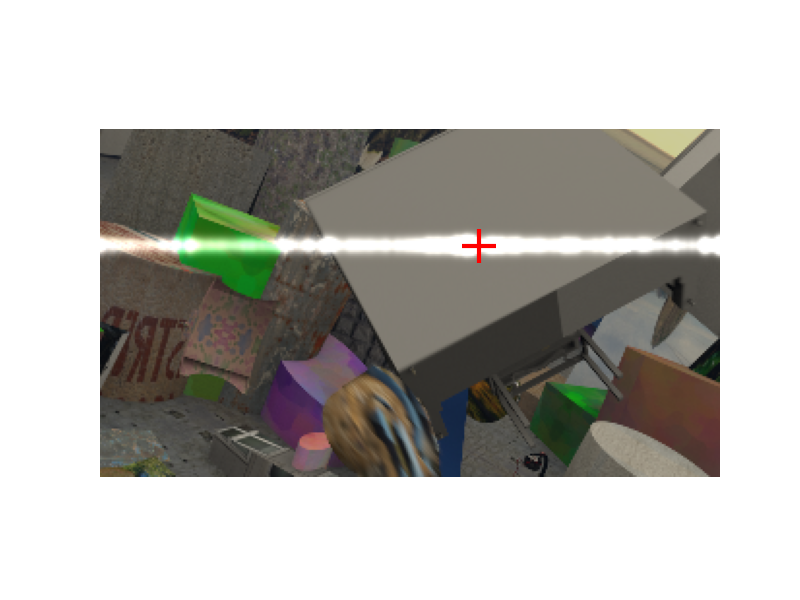}}
    \subfloat{\includegraphics[width=0.16\linewidth,trim={2cm 2cm 2cm 2cm},clip]{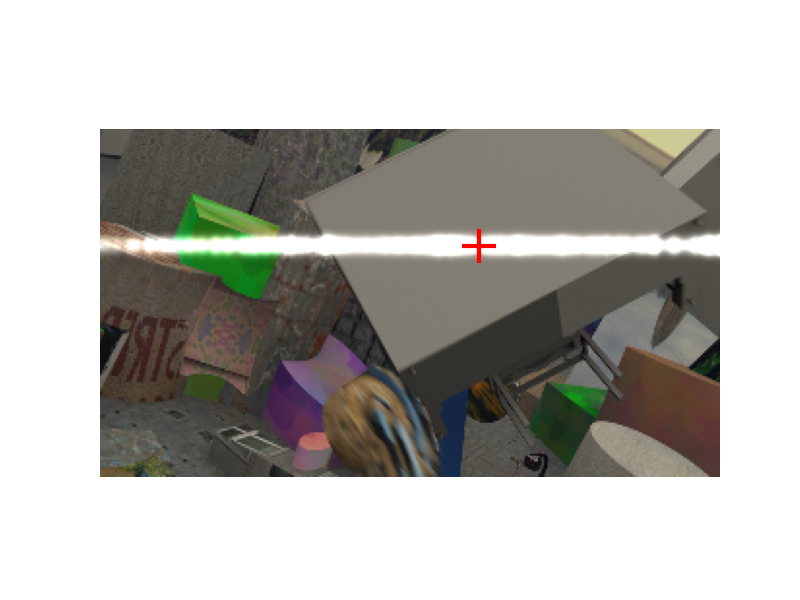}}
    \subfloat{\includegraphics[width=0.16\linewidth,trim={2cm 2cm 2cm 2cm},clip]{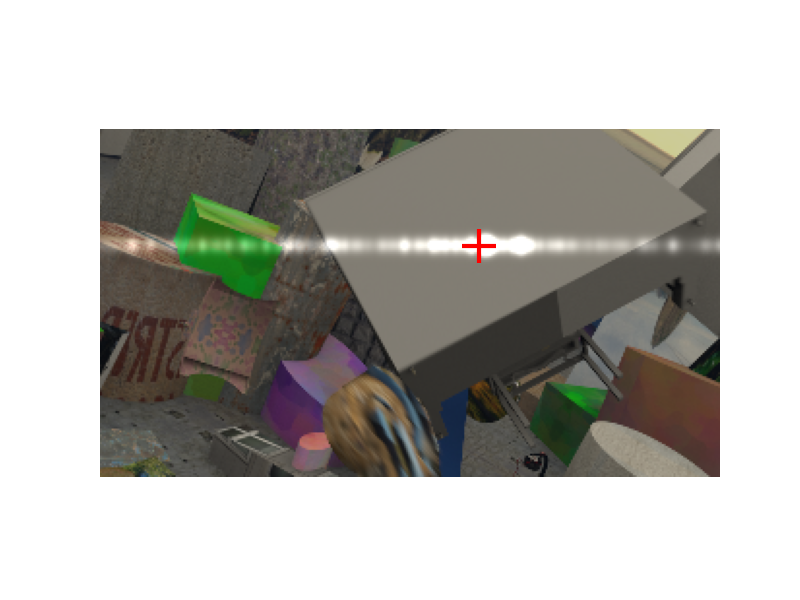}}
    \subfloat{\includegraphics[width=0.16\linewidth,trim={2cm 2cm 2cm 2cm},clip]{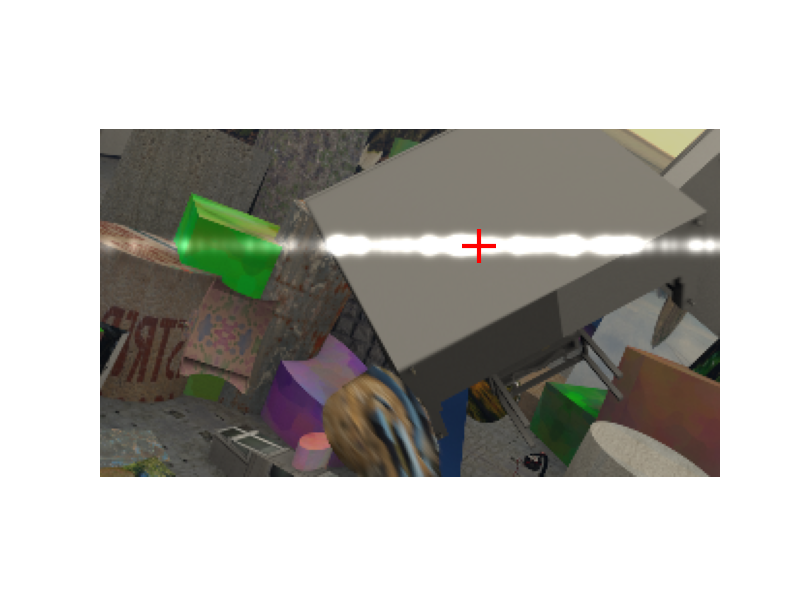}}
    \subfloat{\includegraphics[width=0.16\linewidth,trim={2cm 2cm 2cm 2cm},clip]{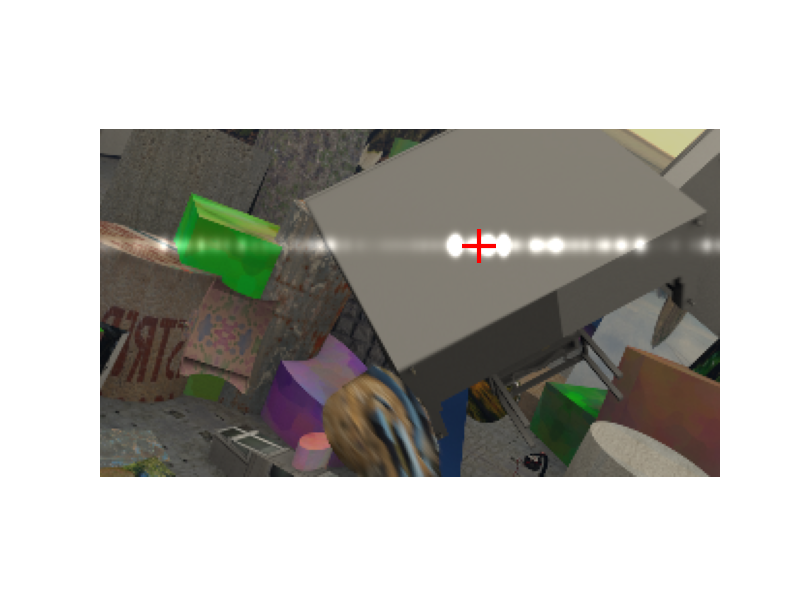}}
    \subfloat{\includegraphics[width=0.16\linewidth,trim={2cm 2cm 2cm 2cm},clip]{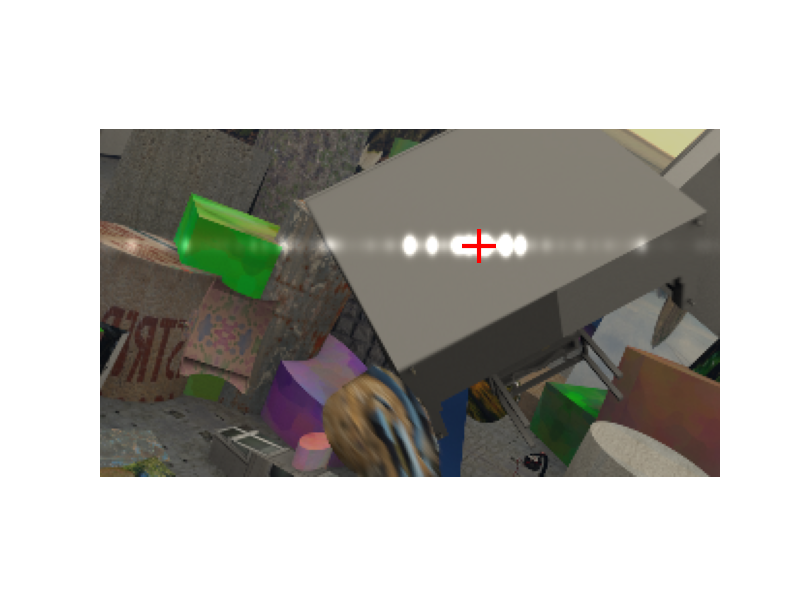}}
    
    \caption{Evolution of self-attention of left image pixel on left image, with source pixel labeled as red cross-hair. First row - attention map of self-attention layers 1-6. Second row - attended pixels of self-attention layers 1-6 are highlighted.}
    \label{fig:evolution_self_attn}
\end{figure*}

\begin{figure*}[htpb]
    \centering
    \subfloat{\includegraphics[width=0.16\linewidth,trim={2cm 2cm 2cm 2cm},clip]{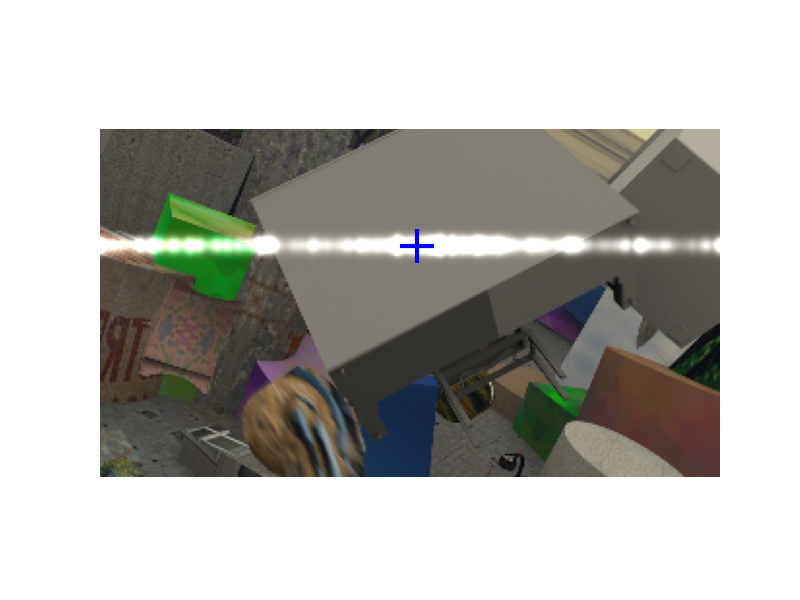}}
    \subfloat{\includegraphics[width=0.16\linewidth,trim={2cm 2cm 2cm 2cm},clip]{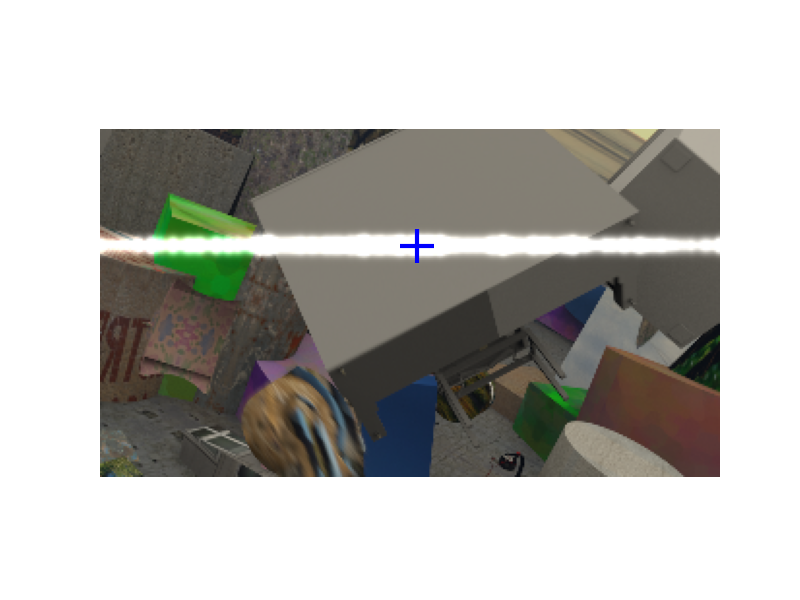}}
    \subfloat{\includegraphics[width=0.16\linewidth,trim={2cm 2cm 2cm 2cm},clip]{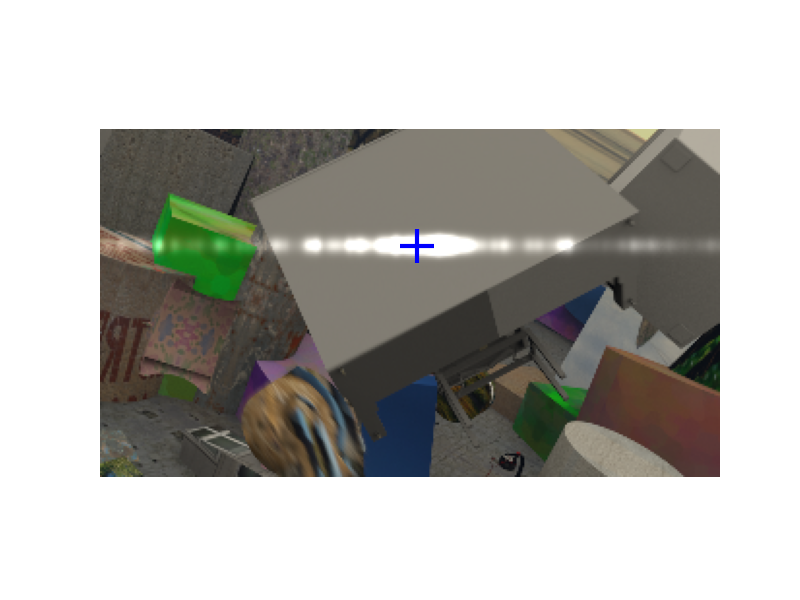}}
    \subfloat{\includegraphics[width=0.16\linewidth,trim={2cm 2cm 2cm 2cm},clip]{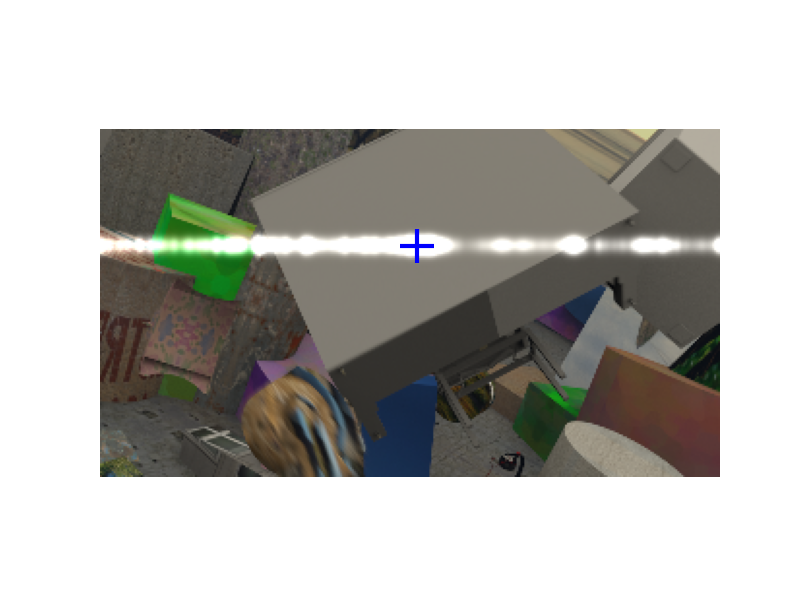}}
    \subfloat{\includegraphics[width=0.16\linewidth,trim={2cm 2cm 2cm 2cm},clip]{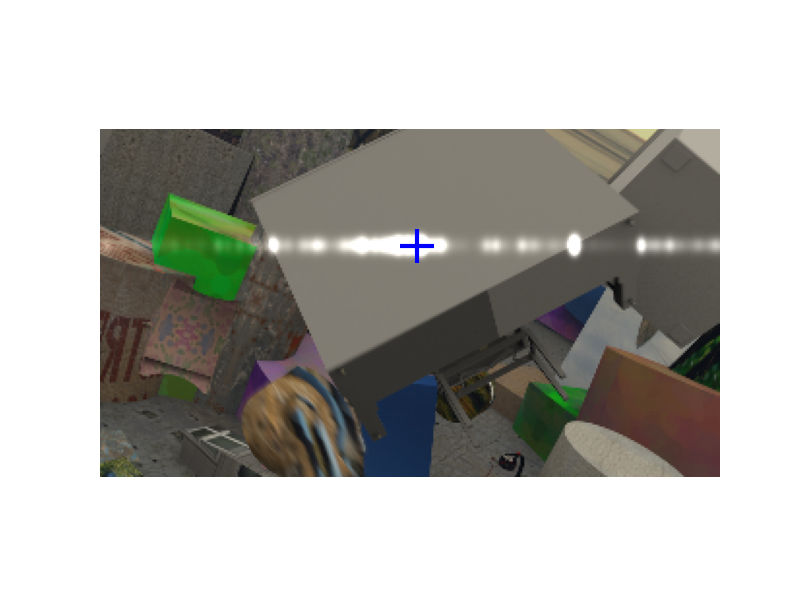}}
    \subfloat{\includegraphics[width=0.16\linewidth,trim={2cm 2cm 2cm 2cm},clip]{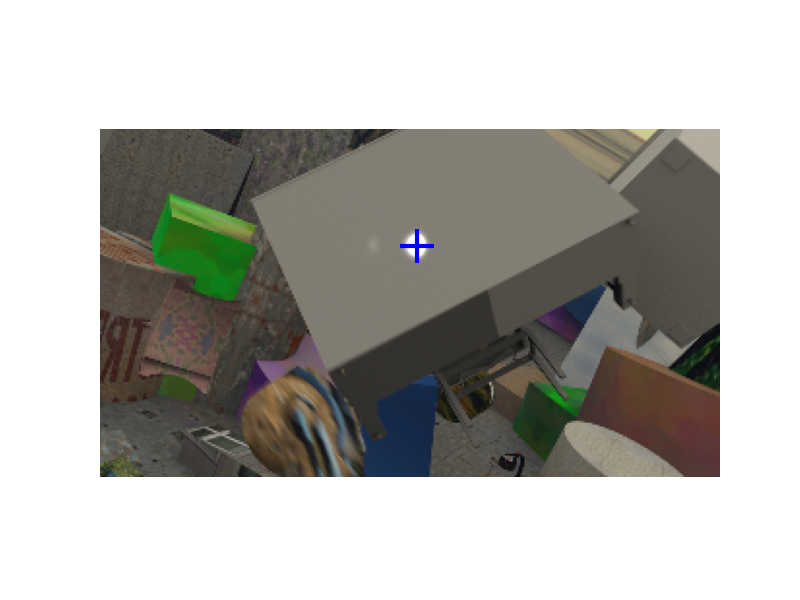}}
    
    \caption{Evolution of cross-attention of source left image pixel on right image, with target ground truth pixel labeled as blue cross-hair. First row - attention map of cross-attention layers 1-6. Second row - attended pixels of cross-attention 1-6 are highlighted.}
    \label{fig:evolution_cross_attn}
\end{figure*}

The evolution of self-attention and cross-attention are also visualized in \autoref{fig:evolution_self_attn} and \autoref{fig:evolution_cross_attn}, where brighter regions are the more attended regions. It can be observed that the attention span from the source pixel (labeled as a red cross-hair) slowly converges to local context as the Transformer progresses, confirming the quantitative result in \autoref{fig:attn_span}. Both attention mechanisms stay focused on edges even if they are far away from the source pixel (i.e., not within the local context). In cross-attention, it can be observed that the final attention shrinks towards the target pixel (labeled as a blue cross-hair).

\vspace{-.3cm}
\section{Generalization Mechanism}
\label{ap:generalization}
\begin{figure}
    \centering
    \subfloat[Scene flow]{\includegraphics[width=0.82\linewidth,trim={8cm 5cm 6cm 2.5cm},clip]{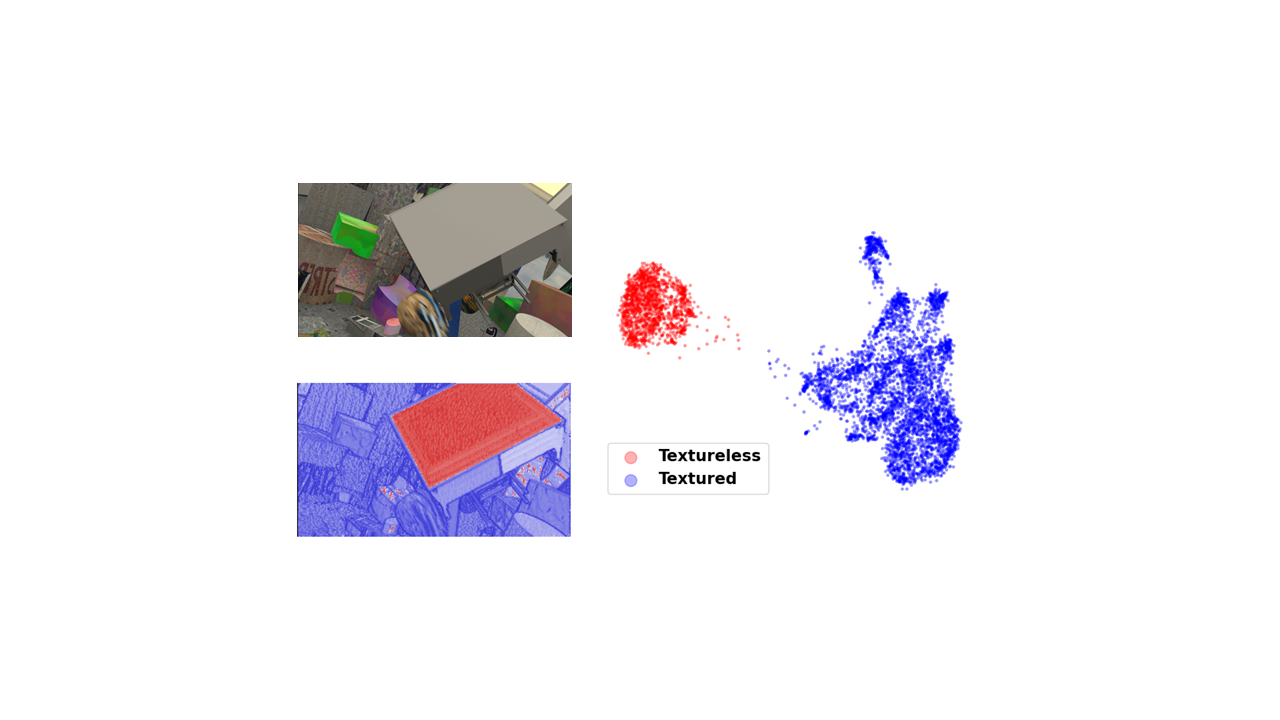}}
     
    \subfloat[All datasets]{\includegraphics[width=0.7\linewidth,trim={0cm 1cm 0cm 1cm},clip]{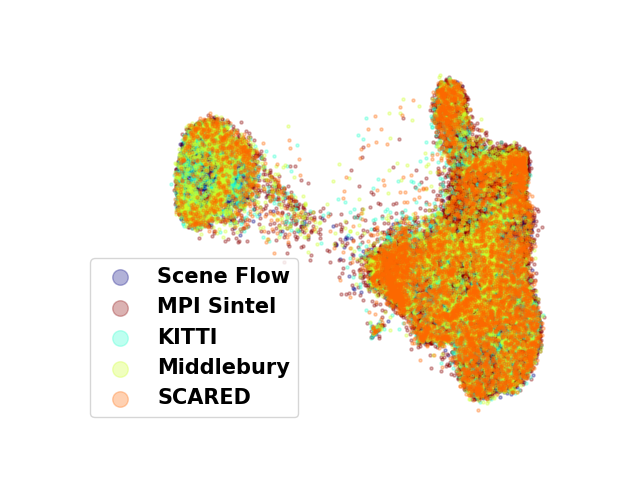}}
    \caption{(a) UMAP visualization of feature map (right) and the corresponding color mask (bottom left) of input image (top left). (b) Umap visualization of all dataset.}
    \label{fig:umap}
\end{figure}

\begin{figure}
    \centering
    \subfloat[MPI Sintel]{\includegraphics[height=0.35\linewidth,trim={8cm 5cm 6cm 5cm},clip]{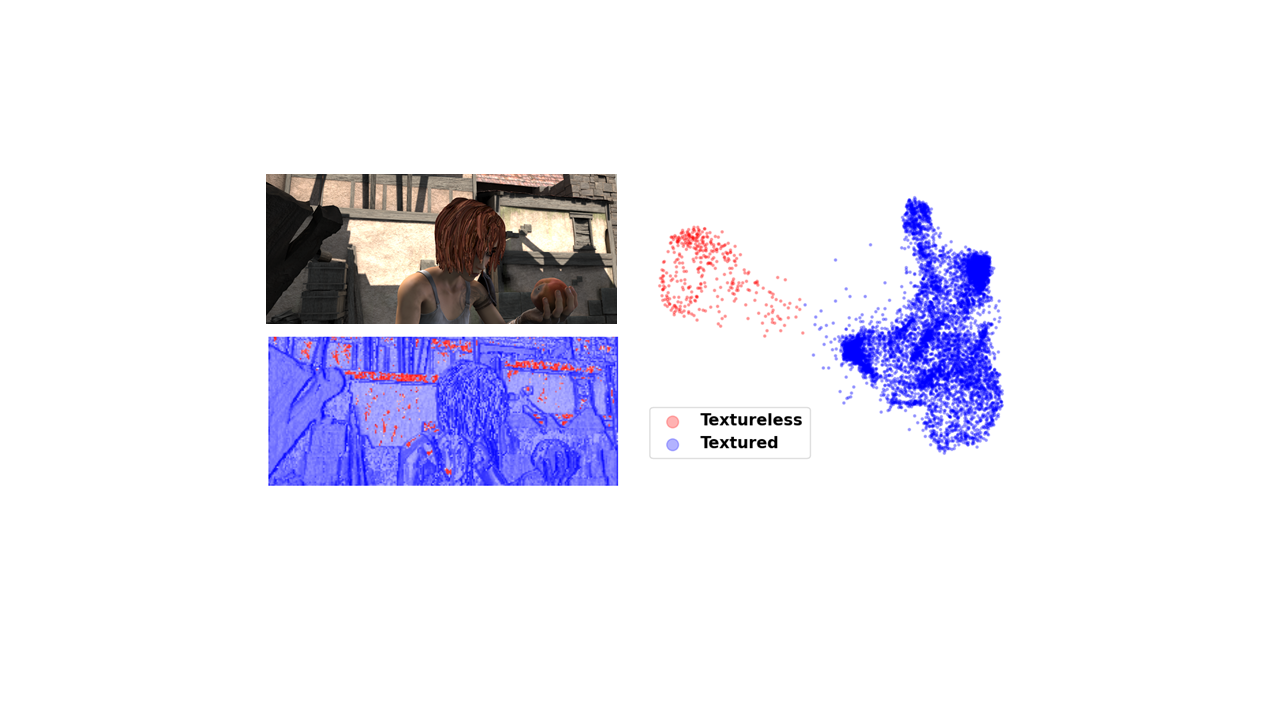}}
    
    \subfloat[KITTI 2015]{\includegraphics[height=0.35\linewidth,trim={6cm 5cm 6cm 4cm},clip]{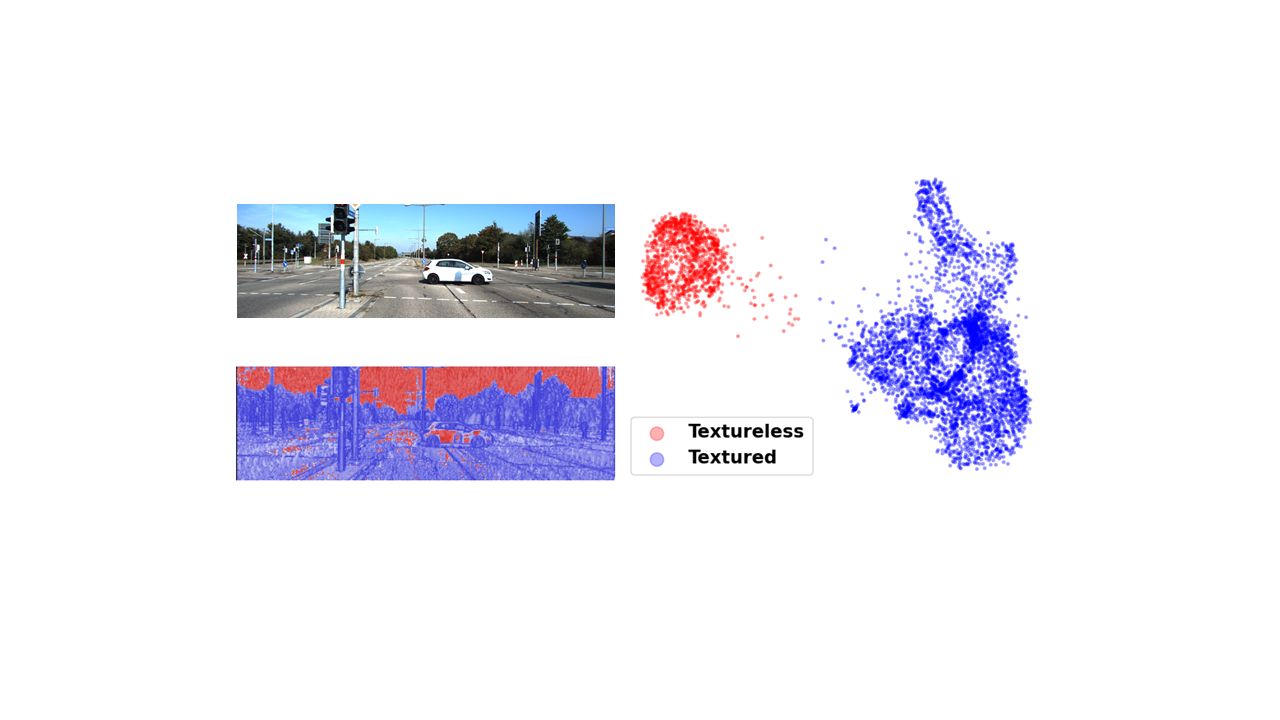}}
    
    \subfloat[Middlebury 2014]{\includegraphics[height=0.5\linewidth,trim={8cm 4cm 8cm 4cm},clip]{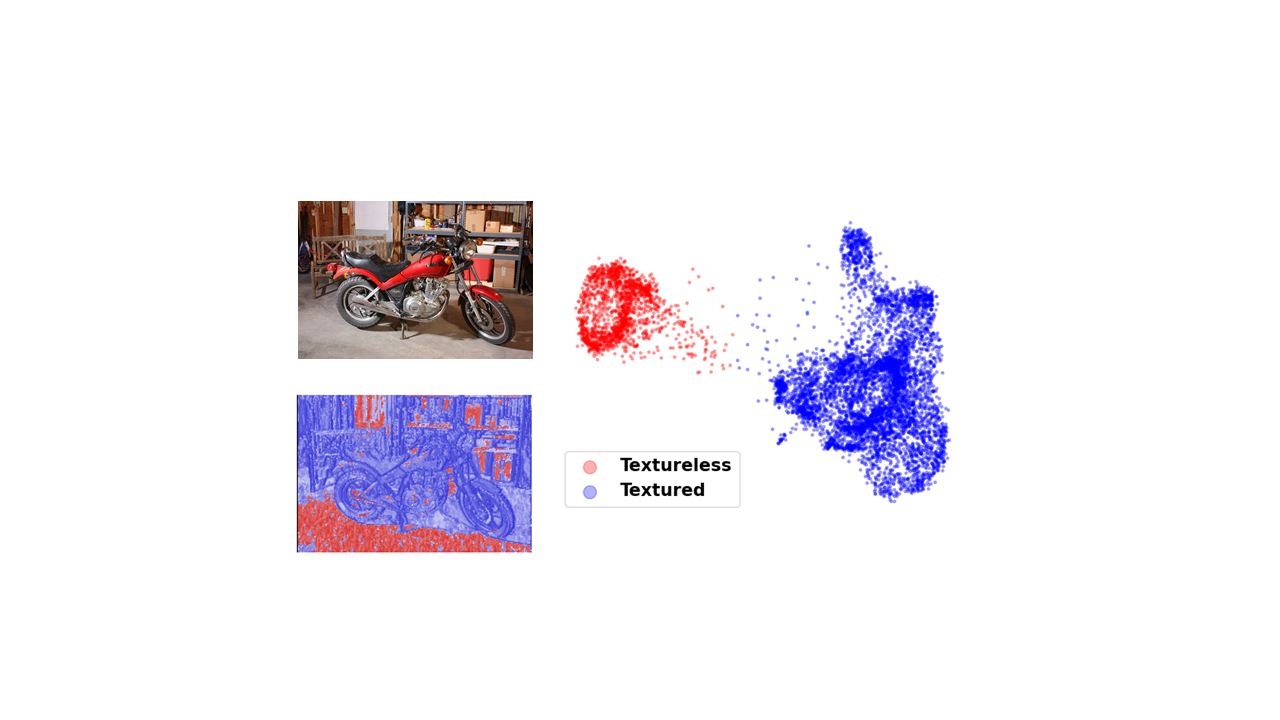}}
    
    \subfloat[SCARED]{\includegraphics[height=0.4\linewidth,trim={8cm 5cm 8cm 5cm},clip]{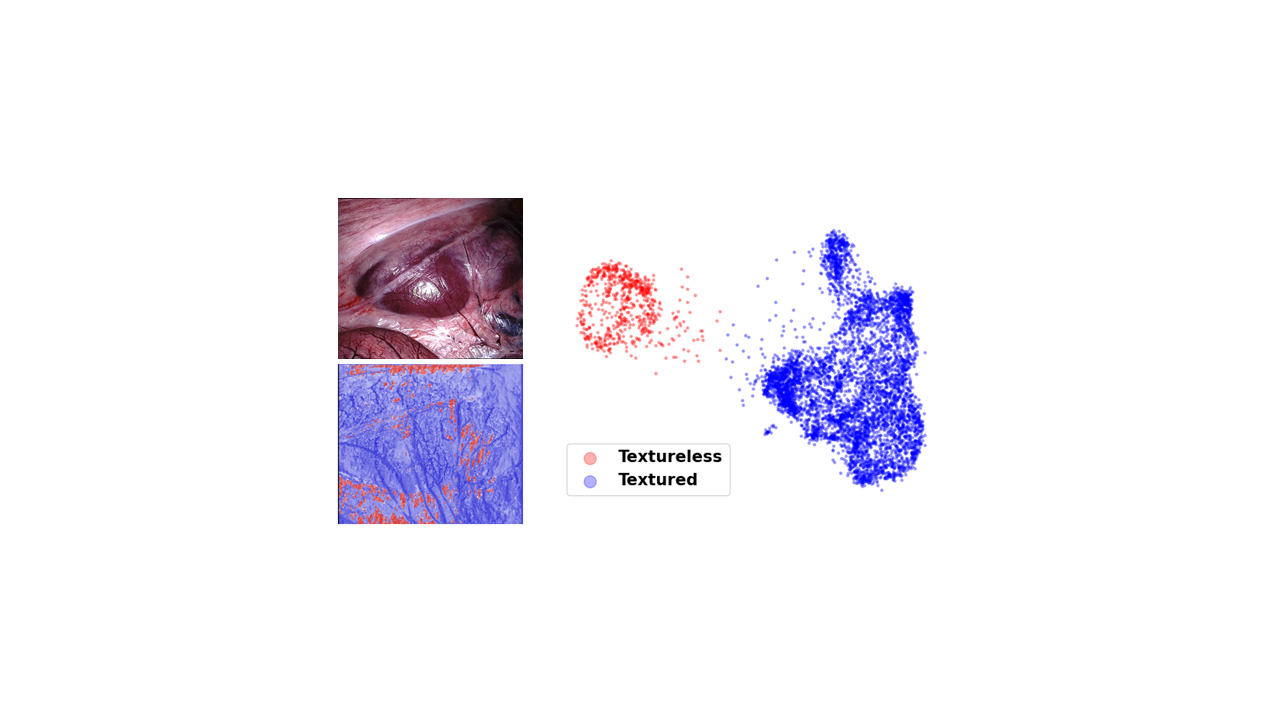}}
    \caption{UMAP visualization of feature map from each domain.}
    \label{fig:umap_vis_individual}
\end{figure}

We visualize the input feature maps to the Transformer using UMAP \cite{mcinnes2018umap} in \autoref{fig:umap}, where the dimensionality reduced embedding is trained only on Scene Flow data. Each data point represents a pixel that the Transformer operates on. We observe that the representations learned by STTR cluster into two regions (\autoref{fig:umap}(a)). To further understand this clustering phenomenon, we visualize the corresponding pixels using a color mask belonging to one of the clusters. The intensity of a pixel in the color mask is higher if it is closer to the centroids of the cluster. Interestingly, the feature extractor groups pixels into textured (blue) and textureless (red) regions. Pixels with a higher intensity in the color mask are mostly correlated to texture edges. To verify that the blue region indeed contains more texture than the red region, we compute the mean Sobel edge-gradient on the normalized image intensities of each dataset. The result is summarized in \autoref{tab:mean_gradient} where, with the exception of SCARED, all datasets have larger magnitude edge-gradients in the blue cluster than red cluster, confirming that blue cluster contains more ``texture'' than red cluster. In \autoref{fig:umap}(b), we show that regardless of the domain, the embeddings are always contained within the same space. We additionally show the individual UMAP reduction of features extracted from MPI Sintel, KITTI 2015, Middlebury 2014, and SCARED dataset on top of Scene Flow in \autoref{fig:umap_vis_individual}(a-d). We hypothesize that this implicitly learnt feature clustering improves the generalization of STTR and makes the Transformer matching process easier.

\begin{table}
    \centering
    \caption{Quantitative comparison of mean edge-gradient based on intensity computed on each dataset in the blue and red clusters.}
    \label{tab:mean_gradient}
    \resizebox{0.7\linewidth}{!}{%
        \begin{tabular}{c|c|c}
            Dataset & Red Cluster& Blue Cluster \\ \hline
            Scene Flow & 0.46 & \textbf{12.02} \\
            MPI Sintel & 16.70 & \textbf{19.15} \\
            KITTI 2015 & 4.06 & \textbf{23.67} \\
            Middleburry 2014 & 7.73 & \textbf{27.48}\\
            SCARED & \textbf{17.39} & 15.42 \\
        \end{tabular}
    }%
\end{table}

\section{Qualitative result of context adjustment layer (CAL)}
\label{ap:cal}
\begin{figure}[htpb]
    \centering
    \subfloat[Disparity \textbf{with} CAL]{\includegraphics[width=0.37\linewidth]{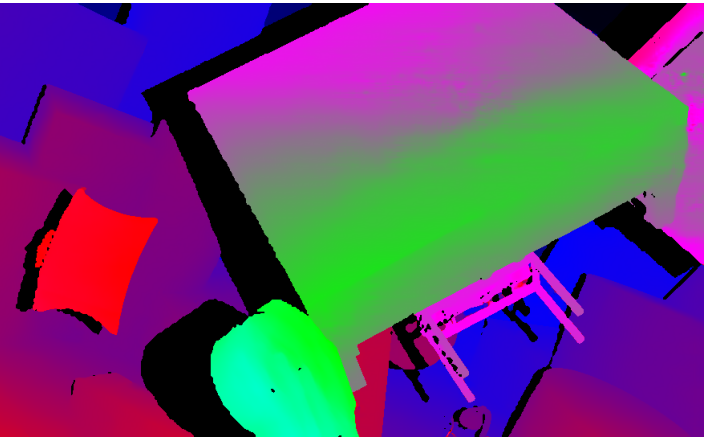}}
    \hspace{0.5em}
    \subfloat[Disparity \textbf{without} CAL]{\includegraphics[width=0.37\linewidth]{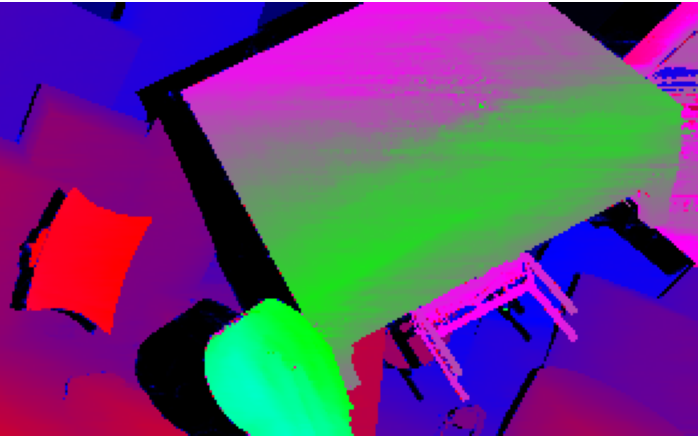}}
    \caption{Qualitative result of disparity with/without CAL.}
    \label{fig:cal}
\end{figure}

We provide visualizations of context adjustment layer's effect in \autoref{fig:cal}. Comparing the CAL output in \autoref{fig:cal}(a), the prediction without CAL in \autoref{fig:cal}(b) the result lacks smoothness due to lack of cross eipolar-line context.

\section{Attention Stride}
\label{ap:attn_stride}
Since the attention module is flexible with respect to the stride of features over which to attend, STTR can run at a faster speed and lower memory footprint at the cost of performance. We do \textit{not} have to re-train STTR as attention stride is only an inference hyperparameter. The result is summarized in \autoref{tab:attn_stride}.

\begin{table}
\centering
\caption{Ablation result on attention stride. Input image resolution is 960$\times$540. Inference performance reported are median \textcolor{red}{memory in GB $\downarrow$} and \textcolor{blue}{speed in Hz $\uparrow$} over 100 runs.}
\label{tab:attn_stride}
\resizebox{0.8\linewidth}{!}{%
    \begin{tabular}{c|c|c|c|c}
    Attention & 3 px &  & Occ  & Inference  \\
    Stride & Error $\downarrow$ & EPE $\downarrow$ & IOU $\uparrow$ & Performance \\ \hline
    3 & \textbf{1.26} & \textbf{0.45} & 0.92 & \textcolor{red}{7.4} / \textcolor{blue}{1.35} \\
    4 & 1.43 & 0.71 & \textbf{0.97} & \textcolor{red}{3.8} / \textcolor{blue}{2.76} \\
    5 & 1.70 & 1.04 & 0.96 & \textcolor{red}{\textbf{2.2}} / \textcolor{blue}{\textbf{4.36}}
    \end{tabular}%
}
\end{table}

\section{Lightweight Implementation}
\label{ap:lightweight}
An additional lightweight model is implemented for STTR for faster inference speed and lighter memory consumption while maintaining the similar performance as STTR. The major changes include:
\begin{itemize}
    \itemsep0em
    \item[--] We remove the flexibility in STTR where attention stride $s$ can change during inference but instead fix it to $s=4$. Thus, the features extracted do not need to be maintained at the full image resolution any more. This reduces memory consumption.
    \item[--] As discussed in Equation 13, the memory consumption is proportional to the number of heads $N_h$. On the other hand, the number of parameters is proportional to $N_h C_h$. Therefore, we can maintain the same number of parameters while halving memory consumption by increasing $C_h$ by two and decreasing $N_h$ by two.
\end{itemize}

We use the same training protocol for the lightweight model as discussed in Sect.~4. We compare the performance of the lightweight STTR and STTR on the Scene Flow benchmark in \autoref{tab:sceneflow_ap} and cross-domain generalization performance on different datasets in \autoref{tab:generalization_ap}. As shown in \autoref{tab:sceneflow_ap}, the performance of lightweight STTR drops compared to STTR in Scene Flow evaluation, especially in terms of 3 px Error and EPE, while inference performance improves with less memory and faster speed. The generalization performance improves in EPE for MPI Sintel, 3\,px Error and EPE for Middlebury 2014 and SCARED, while worsens in 3\,px Error for MPI Sintel and 3\,px error and EPE for KITTI 2015. The occlusion IOU consistently worsens compared to STTR.

\begin{table}[htpb]
    \centering
    \caption{Evaluation on Scene Flow. Inference performance reported are median \textcolor{red}{memory in GB $\downarrow$} and \textcolor{blue}{speed in Hz $\uparrow$} over 100 runs.}
    \resizebox{0.95\linewidth}{!}{%
    \begin{tabular}{c|c|c|c|c}
        & 3\,px & & Occ & Inference\\
        & Error $\downarrow$ & EPE $\downarrow$ & IOU $\uparrow$ & Performance\\ \hline 
        STTR ($s=3$) & \textbf{1.26} & \textbf{0.45} & 0.92 & \textcolor{red}{7.4} / \textcolor{blue}{1.35} \\
        STTR ($s=4$) & 1.43 & 0.71 & \textbf{0.97} & \textcolor{red}{3.8} / \textcolor{blue}{2.76} \\
        Lightweight STTR & 1.54 & 0.50 & \textbf{0.97} & \textcolor{red}{\textbf{2.6}} / \textcolor{blue}{\textbf{5.50}} 
    \end{tabular}
    }%
    \label{tab:sceneflow_ap}
\end{table}

\begin{table*}[htpb]
    \centering
    \caption{Generalization without fine-tuning on MPI Sintel, KITTI 2015, Middlebury 2014, and SCARED dataset. Trained only on Scene Flow dataset.}
    \resizebox{\linewidth}{!}{%
    \begin{tabular}{c||c|c|c||c|c|c||c|c|c||c|c|c}
     & \multicolumn{3}{c||}{MPI Sintel} & \multicolumn{3}{c||}{KITTI 2015} & \multicolumn{3}{c||}{Middlebury 2014} & \multicolumn{3}{c}{SCARED}  \\ \cline{2-13} 
     & 3\,px Error $\downarrow$ & EPE $\downarrow$ & Occ IOU $\uparrow$ & 3\,px Error $\downarrow$ & EPE $\downarrow$ & Occ IOU $\uparrow$ & 3\,px Error $\downarrow$ & EPE $\downarrow$ & Occ IOU $\uparrow$ & 3\,px Error $\downarrow$ & EPE $\downarrow$ & Occ IOU $\uparrow$ \\ \hline 
    STTR & \textbf{5.75} & 3.01 & \textbf{0.86} & \textbf{6.74} & \textbf{1.50} & \textbf{0.98} & 6.19 & 2.33 & \textbf{0.95} & 3.69 & 1.57 & \textbf{0.96} \\
    Lightweight STTR & 5.82 & \textbf{2.95} & 0.69 & 7.20 & 1.56 & 0.95 & \textbf{5.36} & \textbf{2.05} & 0.76 & \textbf{3.30} & \textbf{1.19} & 0.89
    \end{tabular}
    }%
    \label{tab:generalization_ap}
\end{table*}

\section{Inference Memory Consumption and Speed}
\label{ap:inference_stats}
As discussed in Sect.~3.5, STTR can run at a constant memory and speed without a manually limited disparity range. We compare the inference speed and memory consumption with the prior work where a larger disparity range will consume more memory and slow down inference speed. Since PSMNet \cite{chang2018pyramid} uses a feature downsampling rates of 4 and AANet \cite{xu2020aanet} uses a feature pyramid, we set attention stride $s=4$ for STTR to match the setting in PSMNet. As shown in \autoref{tab:inference_mem_speed}, in order for prior work to predict a larger disparity range, the memory consumption increases while inference speed drops. In comparison, STTR runs with a constant memory consumption and speed.

\begin{table}[htpb]
\centering
\caption{Evaluation of inference \textcolor{red}{memory in GB $\downarrow$} and \textcolor{blue}{speed in Hz $\uparrow$} across different disparity ranges. Image resolution (W$\times$H) and maximum disparity values are in pixels. Reported are median values across 100 runs. N/A: disparity range exceeds image width.}
\label{tab:inference_mem_speed}
\resizebox{\linewidth}{!}{%
    \begin{tabular}{c||c||c|c|c|c|c}
    \multirow{2}{*}{Network} & \multirow{2}{*}{\begin{tabular}[c]{@{}c@{}}Image\\ Resolution\end{tabular}} & \multicolumn{5}{c}{Maximum Disparity} \\ \cline{3-7} 
     &  & 192 & 384 & 576 & 768 & 960 \\ \hline
    PSMNet \cite{chang2018pyramid} & \multirow{4}{*}{960 $\times$ 576} & \textcolor{red}{3.9} / \textcolor{blue}{2.21} & \textcolor{red}{7.6} / \textcolor{blue}{1.20} & \textcolor{red}{11.4} / \textcolor{blue}{0.85} & \textcolor{red}{15.1} / \textcolor{blue}{0.62} & \textcolor{red}{18.8} / \textcolor{blue}{0.50} \\ \cline{3-7} 
    AANet \cite{xu2020aanet} &  & \textcolor{red}{0.6} / \textcolor{blue}{14.01} & \textcolor{red}{0.7} / \textcolor{blue}{9.65} & \textcolor{red}{0.9} / \textcolor{blue}{6.91} & \textcolor{red}{1.1} / \textcolor{blue}{5.59} & \textcolor{red}{1.3} / \textcolor{blue}{4.28} \\ \cline{3-7} 
    \textbf{STTR} &  & \multicolumn{5}{c}{\textcolor{red}{3.8} / \textcolor{blue}{2.76}} \\
    \textbf{Lightweight STTR} &  & \multicolumn{5}{c}{\textcolor{red}{2.0} / \textcolor{blue}{4.42}} \\ \hline
    PSMNet \cite{chang2018pyramid} & \multirow{4}{*}{672 $\times$ 480} & \textcolor{red}{2.3} / \textcolor{blue}{3.89} & \textcolor{red}{4.5} / \textcolor{blue}{2.00} & \textcolor{red}{6.6} / \textcolor{blue}{1.39} & \multicolumn{2}{c}{N/A} \\ \cline{3-7} 
    AANet \cite{xu2020aanet} & & \textcolor{red}{0.3} / \textcolor{blue}{20.14} & \textcolor{red}{0.4} / \textcolor{blue}{14.70} & \textcolor{red}{0.5} / \textcolor{blue}{11.06} & \multicolumn{2}{c}{N/A} \\ \cline{3-7} 
    \textbf{STTR} &  & \multicolumn{3}{c|}{\textcolor{red}{1.8} / \textcolor{blue}{5.77}} & \multicolumn{2}{c}{N/A} \\
    \textbf{Lightweight STTR} &  & \multicolumn{3}{c|}{\textcolor{red}{0.87} / \textcolor{blue}{9.11}} & \multicolumn{2}{c}{N/A} \\ \hline
    PSMNet \cite{chang2018pyramid} & \multirow{4}{*}{384 $\times$ 192} & \multicolumn{2}{c|}{Resolution Too Small} & \multicolumn{3}{c}{N/A} \\ \cline{3-7} 
    AANet \cite{xu2020aanet} & & \textcolor{red}{0.1} / \textcolor{blue}{22.1} & \textcolor{red}{0.1} / \textcolor{blue}{21.4} & \multicolumn{3}{c}{N/A} \\ \cline{3-7} 
    \textbf{STTR} &  & \multicolumn{2}{c|}{\textcolor{red}{0.3} / \textcolor{blue}{18.9}} & \multicolumn{3}{c}{N/A} \\
    \textbf{Lightweight STTR} &  & \multicolumn{2}{c|}{\textcolor{red}{0.2} / \textcolor{blue}{25.6}} & \multicolumn{3}{c}{N/A} \\ \hline
    \end{tabular}%
}
\end{table}

\section{Dataset Information and Pre-processing}
\label{ap:dataset_info}
Scene Flow \cite{mayer2016large} FlyingThings3D Full dataset (final pass) provides realistic artifacts but does not provide occlusion information. Therefore, we subsample the Full dataset using the corresponding occlusion information from DispNet/FlowNet2.0 dataset. After pre-processing, Scene Flow contains 21818 training images of resolution 960$\times$540, with maximum disparity 602 px (0.67 of image width). The test dataset contains 4248 images with maximum disparity 468 (0.49 of image width). MPI Sintel \cite{butler2012naturalistic} contains 1063 images of resolution $1024\times436$, with maximum disparity 487 px (0.46 of image width). KITTI 2015 \cite{menze2015object} contains 200 images of resolution $1242\times375$, with maximum disparity of 192 px (0.15 of image width). We note that pixels with disparities larger than 192 are intentionally masked out in the dataset. Middlebury 2014 \cite{scharstein2014high} quarter resolution subset contains 15 images of various image resolution, with maximum disparity of 161 px (0.22 of image width). SCARED \cite{allan2021stereo} requires additional pre-processing since it only provides the depth data and corresponding camera intrinsics parameters. There are 7 subsets in total, containing 27 videos. Since subsets \{4,5\} contain very large camera intrinsic errors \cite{allan2021stereo}, we choose to leave them out of our evaluation since this introduces unnecessary uncertainty. Furthermore, other than the first frames of each video, subsequent frames are interpolated using the kinematics information of the robot with synchronization and kinematics error. Therefore, the depth values for subsequent frames are not accurate. We also exclude those images for reliable evaluation. The left and right 100 pixels were cropped due to invalidity after rectification. After pre-processing, SCARED contains 19 images of resolution $1080\times1024$ with maximum disparity of 263 px (0.24 of image width).

\section{Loose Analogy to Biological Stereo Vision}
It has been shown that the biological stereo vision system (e.\,g. that of a human) perceives depth from stereo images by relying on low-level cortex cues. One example that demonstrates this effect is the random-dot stereograms experiment \cite{julesz1971foundations}, where there is no meaningful texture in the images but only random dots; yet, humans can still perceive the 3D objects. At the same time, the biological vision system imposes geometric assumptions on objects as a piece-wise smoothness prior \cite{furukawa2009manhattan}, which involves mid-level cortex processing. In a way, STTR emulates the biological stereo system in that the Transformer processes the images at a low-level to finds matches between features. STTR then locally refines the raw disparity using the context adjustment layer, which is in loose analogy to mid-level vision.

\section{Training Time and Number of parameters}
\label{ap:training_param}
The total training time on Scene Flow dataset \cite{mayer2016large} for STTR is approximately 120 hours on one Titan RTX GPU with batch size 1 for 15 epochs. We note that the training time will vary with the batch size and number/type of GPUs used. The number of parameters of STTR compared with contemporary architectures is summarized in \autoref{tab:num_of_param}. Other than  LEAStereo \cite{cheng2020hierarchical} which is optimized using Neural Architecture Search, STTR has the least number of parameters.

\begin{table}[htpb]
    \centering
    \caption{Number of parameters in contemporary architectures for stereo depth estimation.}
    \resizebox{0.7\linewidth}{!}{%
    \begin{tabular}{c|c|c}
        \textbf{Approach} & \textbf{Network} & \textbf{Params} {[}M{]} \\ \hline
        \multirow{4}{*}{3D Convolution} 
        & GANet-11 \cite{zhang2019ga} & 6.6 \\
        & PSMNet \cite{chang2018pyramid} & 5.2 \\
        & GC-Net\cite{kendall2017end} & 3.5 \\ & LEAStereo \cite{cheng2020hierarchical} & \textbf{1.8} \\
        \hline
        
        \multirow{4}{*}{Correlation} & iResNet \cite{liang2018learning} & 43 \\
        & DispNetCorr1D \cite{mayer2016large} & 42 \\ 
        & HD$^3$ \cite{yin2019hierarchical} &  39 \\
        & AANet \cite{xu2020aanet} &  3.9 \\
        \hline
        
        Hybrid & GwcNet-g \cite{guo2019group} & 6.5 \\ \hline
        Classification & Bi3D \cite{badki2020bi3d} & 37 \\ \hline
        \textbf{Transformer} & \textbf{STTR (Ours)} & 2.5    \end{tabular}
    }
    \label{tab:num_of_param}
\end{table}

{\small
\bibliographystyle{ieee_fullname}
\bibliography{egbib}
}

\end{document}